\documentclass[twoside]{article}

\usepackage[accepted]{aistats2023}
\usepackage[round]{natbib}

\usepackage{amsmath}
\usepackage{amssymb}
\usepackage{mathtools}
\usepackage{amsthm}
\usepackage{multirow}
\usepackage{subcaption}
\usepackage{algorithm}
\usepackage{algorithmic}
\usepackage{booktabs}
\usepackage{units}

\newtheorem{theorem}{Theorem}[section]
\newtheorem{proposition}[theorem]{Proposition}

\newcommand{\eg}{e.\,g., }
\newcommand{\ie}{i.\,e., }

\newcommand{\oms}{\{\!\!\{}
\newcommand{\cms}{\}\!\!\}}

\begin{document}

%
\runningtitle{Published as a conference paper at AISTATS 2023}

%

\twocolumn[

\aistatstitle{Weisfeiler and Leman go Hyperbolic: Learning \\Distance Preserving Node Representations}

\aistatsauthor{ Giannis Nikolentzos \And Michail Chatzianastasis \And  Michalis Vazirgiannis }

\aistatsaddress{ LIX, \'Ecole Polytechnique \\ IP Paris, France \And  LIX, \'Ecole Polytechnique \\ IP Paris, France \And LIX, \'Ecole Polytechnique \\ IP Paris, France } ]

\begin{abstract}
In recent years, graph neural networks (GNNs) have emerged as a promising tool for solving machine learning problems on graphs.
Most GNNs are members of the family of message passing neural networks (MPNNs).
There is a close connection between these models and the Weisfeiler-Leman (WL) test of isomorphism, an algorithm that can successfully test isomorphism for a broad class of graphs.
Recently, much research has focused on measuring the expressive power of GNNs.
For instance, it has been shown that standard MPNNs are at most as powerful as WL in terms of distinguishing non-isomorphic graphs.
However, these studies have largely ignored the distances between the representations of nodes/graphs which are of paramount importance for learning tasks.
In this paper, we define a distance function between nodes that is based on the hierarchy produced by the WL algorithm and propose a model that learns representations which preserve those distances between nodes.
Since the emerging hierarchy corresponds to a tree, to learn these representations, we capitalize on recent advances in the field of hyperbolic neural networks.
We empirically evaluate the proposed model on standard node and graph classification datasets where it achieves competitive performance with state-of-the-art models.
\end{abstract}

\section{Introduction}
Over the past few years, graph neural networks (GNNs) have been applied with great success to machine learning problems on graphs in various application domains.
For instance, in chemistry, much attention has been devoted to deep learning systems for drug screening or design where molecules are represented as graphs~\citep{kearnes2016molecular}.
Likewise, in biology, an issue of high interest is the prediction of the functions of proteins modeled as graphs~\citep{you2021deepgraphgo}.
So far, the field of GNNs has been largely dominated by message passing architectures.
Indeed, most of these models share the same basic idea and can be reformulated into a single common framework, so-called message passing neural networks (MPNNs)~\citep{gilmer2017neural}.
Specifically, these models follow a message passing procedure, where each node updates its feature vector by aggregating the feature vectors of its neighbors.
This iterative scheme runs for a number of steps, and then to compute a feature vector for the entire graph, MPNNs typically employ some permutation invariant readout function such as summing or averaging the feature vectors of all the nodes of the graph.

Recently, much research has focused on measuring the expressive power and limitations of MPNNs.
One line of research focused on establishing a connection between MPNNs and the Weisfeiler-Leman (WL) test of graph isomorphism~\citep{weisfeiler1968reduction}, a powerful heuristic that can successfully test isomorphism for a broad class of graphs~\citep{babai1979canonical}.
It was shown that standard MPNNs do not have more power in terms of distinguishing non-isomorphic graphs than the WL algorithm~\citep{morris2019weisfeiler,xu2019powerful}.
Based on those findings, in the past years, considerable effort was devoted to the development of models that are more powerful than the WL algorithm~\citep{morris2019weisfeiler,maron2019provably,chen2019equivalence,morris2020weisfeiler}.
Most of the above studies investigate the power of GNNs in terms of distinguishing non-isomorphic graphs.
However, in graph classification/regression problems, we are not that much interested in testing whether two (sub)graphs are isomorphic to each other, and it has been observed that stronger GNNs (in the aforementioned sense) do not necessarily outperform weaker GNNs \citep{dwivedi2020benchmarking}.
On the other hand, we would like to learn representations which could preserve the similarities or distances of subgraphs/graphs.
Deep neural networks are known for being very sensitive to their input.
For instance, it has been reported that GNNs are extremely vulnerable to adversarial attacks on the graph structure~\citep{dai2018adversarial}.
Such adversarial attacks can mislead a GNN and significantly decrease its performance.
To increase the robustness of standard neural network layers, previous studies proposed to constrain their local Lipschitz constants, \ie to make their output preserve distances as much as possible for some given distance function~\citep{virmaux2018lipschitz}.
Furthermore, functions that vary at a slower rate are usually considered simpler and generalize better~\citep{gouk2021regularisation}.
Thus, in many settings, it is of paramount importance to learn representations that preserve the distances of nodes.

\noindent\textbf{Present Work.}
To achieve that, in this paper, we propose a new MPNN that learns node representations that respect the distances between nodes, as those are defined by the WL algorithm.
The WL algorithm iteratively updates a given node's color (or label) by aggregating the colors of its neighbors.
The color classes produced in a given iteration of the algorithm are at least as fine as those produced in the previous iteration.
Thus, the sequence of node colors gives rise to a family of nested subsets, which can naturally be represented by a hierarchy (\ie a tree).
It is well-known that trees can be embedded with arbitrarily low distortion into the hyperbolic space, while Euclidean space cannot achieve such a low distortion \citep{sarkar2011low}.
Therefore, Euclidean MPNNs cannot encode accurately the information contained in the WL hierarchy.
Thus, in this paper, we propose a MPNN which delivers the ``best of both worlds'' from Euclidean space and hyperbolic space.
Our proposed model, so-called \textit{Weisfeiler-Leman Hyperbolic Network} (WLHN), takes into account the distance of the input nodes according to the hierarchy induced by WL, but also according to the representations that emerge from the neighborhood aggregation procedure.
To embed the WL tree hierarchy into the hyperbolic space, we capitalize on recent advances in hyperbolic representations using the Poincaré ball model \citep{ganea2018hyperbolic,pmlr-v139-chami21a}. 
Specifically, we propose an embedding algorithm (DiffHypCon) which generalizes Sarkar's algorithm \citep{sarkar2011low} such that the model is differentiable and end-to-end trainable.
Experiments on standard benchmark datasets demonstrate that the proposed model either outperforms or achieves performance comparable with that of state-of-the-art models in node and graph classification tasks. 


\section{Related Work}\label{sec:related_work}

\noindent\textbf{GNNs.}
The first GNN models were proposed several years ago~\citep{sperduti1997supervised,scarselli2009graph,micheli2009neural}.
However, it was not until recently that this family of models attracted significant attention, mainly due to the advent of deep learning~\citep{bruna2014spectral,li2015gated,defferrard2016convolutional,kipf2017semi,hamilton2017inductive,zhang2018end}.
Broadly speaking, these models can be categorized into spectral and spatial approaches depending on which domain the convolutions (neighborhood aggregations) are applied to.
Interestingly, the majority of these models follow the same principle and can be reformulated into a single common framework.
These models are known as Message Passing Neural Networks (MPNNs)~\citep{gilmer2017neural}.
These models employ an iterative message passing procedure, where each node combines the representations of its neighbors with its own representation to compute a new representation.
This procedure is typically followed by a readout (or pooling) phase where a feature vector for the entire graph is produced using some permutation invariant function.
Several works have proposed extensions and improvements to the message passing procedure of MPNNs.
For instance, some works have proposed more expressive or learnable aggregation functions \citep{murphy2019relational,seo2019discriminative,dasoulas2021learning,chatzianastasis2022graph}, schemes that incorporate different local structures or high-order neighborhoods \citep{abu2019mixhop,jin2020gralsp,nikolentzos2020k}, approaches that utilize node positional information \citep{you2019position}, while others have focused on efficiency \citep{gallicchio2020fast}. 
Fewer works have focused on the readout phase and have proposed more sophisticated pooling functions~\citep{such2017robust,gao2019graph}.
Note that there also exist GNNs that are not MPNNs~\citep{nikolentzos2022permute}.

\noindent\textbf{Expressive power of GNNs.}
Recently, a line of research has started exploring the expressive power of GNNs.
Several of those studies have investigated how GNNs are related to the WL test of isomorphism and its higher-order variants.
For instance, it was shown that standard GNNs are at most as powerful as the WL algorithm in terms of distinguishing non-isomorphic graphs~\citep{morris2019weisfeiler,xu2019powerful}.
Other studies proposed families of GNNs whose message passing scheme is equivalent to high-order variants of the WL algorithm, and can thus distinguish more pairs of non-isomorphic graphs than standard MPNNs~\citep{morris2019weisfeiler,morris2020weisfeiler}.
\cite{maron2019provably} introduced a class of GNNs which are at least as powerful as the folklore variant of the $k$-WL graph isomorphism test in terms of distinguishing non-isomorphic graphs, while \cite{chen2019equivalence} also proposed a GNN that is more powerful than the $2$-WL algorithm. 
\cite{barcelo2020logical} characterized the expressive power of GNNs in terms of classical logical languages based on a connection between first-order logic and the WL algorithm.
\cite{sato2021random} showed that random features make standard MPNNs more expressive. 
Some works propose neighborhood aggregation schemes that take into account all possible node permutations and produce universal graph representations~\citep{murphy2019relational,dasoulas2020coloring}.
However, due to the dramatically high complexity, approximation schemes are necessary.
For a comprehensive overview of the expressive power of GNNs, the interested reader is referred to the survey by~\cite{sato2020survey}.
The above studies investigate the power of GNNs in terms of distinguishing non-isomorphic graphs or in terms of how well they can approximate combinatorial problems.
On the other hand, the proposed model learns representations that capture the distance of nodes which is largely ignored by the above models.

\noindent\textbf{Hyperbolic Embedding Algorithms and Neural Networks.}
In the past years, there has been a growing interest in algorithms that can learn hyperbolic embeddings.
Such algorithms can be very effective in embedding hierarchical graphs.
\cite{nickel2017poincare} proposed a new algorithm for learning hierarchical representations of symbolic data by embedding them into the hyperbolic space using the Poincar{\'e}-ball model.
Later, they proposed a new optimization approach based on the Lorentz model of hyperbolic space for learning, while they found that learning embeddings in the Lorentz model are more efficient than in the Poincar{\'e}-ball model~\citep{nickel2018learning}.
\cite{ganea2018entailment} proposed an algorithm for embedding directed acyclic graphs in the Poincar{\'e}-ball model, while \cite{balazevic2019multi} proposed an algorithm for embedding hierarchical multi-relational data also in the Poincar{\'e}-ball model.
\cite{chami2020low} introduced a class of hyperbolic embedding models that can capture both hierarchical and logical patterns.
Some other studies have extended deep learning methods to the hyperbolic space by deriving hyperbolic versions of common neural network layers such as recurrent layers and feed-forward layers~\citep{ganea2018hyperbolic,shimizu2020hyperbolic}.
The works closest to ours are the ones proposed by~\cite{chami2019hyperbolic} and by~\cite{liu2019hyperbolic}, where the authors generalize Euclidean MPNNs to the hyperbolic space.
However, both these models assume that there exists a hierarchical structure in the input data, and they aim to capture such latent hierarchies.
On the other hand, in our setting, we have access to an explicit hierarchy produced by the WL algorithm and our model learns representations that respect that hierarchy.

\section{Preliminaries}\label{sec:preliminaries}

\noindent\textbf{Notation.}
Let $\mathbb{N}$ denote the set of natural numbers, \ie $\{1,2,\ldots\}$. 
Then, $[n] = \{1,\ldots,n\} \subset \mathbb{N}$ for $n \geq 1$.
Let also $\oms \cms$ denote a multiset, \ie a generalized concept of a set that allows multiple instances for its elements.
Let $G = (V,E)$ be an undirected graph, where $V$ is the vertex set and $E$ is the edge set.
We will denote by $n$ the number of vertices and by $m$ the number of edges, \ie $n = |V|$ and $m = |E|$.
Let $\mathcal{N}(v)$ denote the the neighborhood of vertex $v$, \ie the set $\{u \mid \{v,u\} \in E\}$.
The degree of a vertex $v$ is $\deg(v) = |\mathcal{N}(v)|$.
Two graphs $G = (V,E)$ and $G' = (V',E')$ are isomorphic (denoted by $G \cong G'$) if there is a bijective mapping $f : V \rightarrow V'$ such that $(v,u) \in E$ iff $(f(v),f(u)) \in E'$.

\begin{figure*}[t]
 \centering
 \begin{subfigure}[c]{0.5\textwidth}
     \centering
     \includegraphics[width=.7\textwidth]{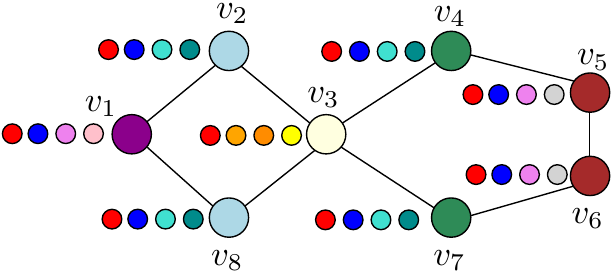}
     \caption{A graph $G$}
 \end{subfigure}%
 \begin{subfigure}[c]{0.5\textwidth}
     \centering
     \includegraphics[width=.7\textwidth]{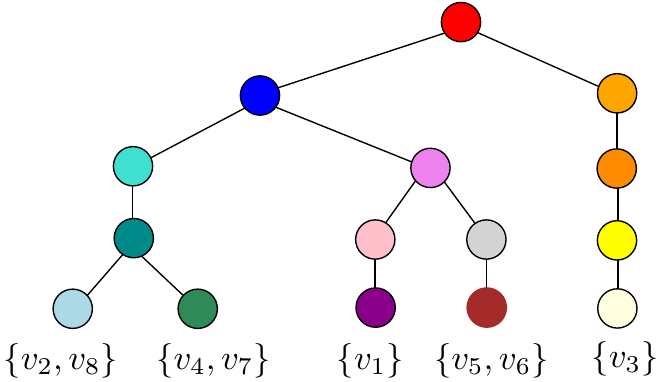}
     \caption{hierarchy $H_G$}
 \end{subfigure}%
    \caption{An illustration of (a) a graph $G$ with uniform initial colors $c_0$ and refined colors $c_i$ for $i \in [4]$, and of (b) its corresponding WL tree hierarchy $H_G$. The nodes of $G$ are the leaves of the hierarchy.}
    \label{fig:example}
\end{figure*}

\noindent\textbf{WL Algorithm.}
The WL algorithm (also known as color refinement) has been revisited a lot in the past years since it found its way into several machine learning applications~\citep{morris2021weisfeiler}.
The algorithm iteratively computes a coloring of the set of nodes of a graph.
Formally, given a graph $G=(V,E)$, a coloring of the set of nodes $V$ is a mapping $c \colon V \rightarrow \mathbb{N}$.
In other words, a coloring $c$ assigns a number (or color) to every node of the graph.
The WL algorithm runs for a number of iterations and is associated with a sequence of colorings $c_0, c_1, \ldots, c_T$.
If the initial coloring $c_0$ of $V$ is not specified, we either assume a monochromatic coloring, \ie all vertices have the same color or we define a coloring that maps nodes to their degrees, \ie $c_0(v) = \text{deg}(v)$.
The refinement of a coloring $c_i$ is a new coloring $c_{i+1}$ defined as follows.
For every vertex $v$, collect and lexicographically sort the multiset of colors of its neighbors $\oms {c_i(u) | u \in \mathcal{N}(v)} \cms$.
Let $\mathcal{M}_i(v)$ denote the sequence of colors that emerges from the above operation.
Then, define $(c_i(v), \mathcal{M}_i(v))$ and assign a new color $c_{i+1}(v)$ to node $v$ by employing a one-to-one mapping from tuples such as $(c_i(v), \mathcal{M}_i(v))$ to new colors.
The new color $c_{i+1}(v)$ of $v$ depends on the colors of its neighbors and the previous color of $v$.
Thus, in iteration $i+1$, vertices $v$ and $u$ receive different colors, if they already had different colors in iteration $i$, or if the multisets of colors of their neighbors in iteration $i$
are different.
Observe also that each coloring $c_0, c_1, \ldots, c_T$ partitions $V$ into color classes, \ie sets of nodes with the same color.
Since the implication $c_i(v) \neq c_i(u) \Rightarrow c_{i+1}(v) \neq c_{i+1}(u)$ holds for all $v, u \in V$ and any $i \in \mathbb{N}$, the color classes produced in iteration $i+1$ are at least as fine as those produced in iteration $i$.
Hence, the sequence of colorings $c_0, c_1, \ldots, c_T$ gives rise to a family of nested subsets, which can naturally be represented by a hierarchy.
Figure~\ref{fig:example} illustrates the color refinement procedure and the hierarchy that emerges from that procedure.

\noindent\textbf{Poincar{\'e} Ball Model.}
A hyperbolic space is a non-Euclidean space with constant negative curvature.
Specifically, the Poincar{\'e} ball model of hyperbolic space corresponds to the Riemannian manifold $(\mathbb{D}^n, g_\mathbf{x}^\mathbb{D})$ where $\mathbb{D}^n = \{ \mathbf{x} \in \mathbb{R}^n \mid ||\mathbf{x}|| < 1 \}$, \ie an open unit ball.
A Riemannian manifold is a real and smooth manifold equipped with an inner product which is called a Riemannian metric.
The above manifold is equipped with the metric $g_\mathbf{x}^\mathbb{D} = \lambda_\mathbf{x}^2 g^E$ where $\lambda_\mathbf{x} = \frac{2}{1-||\mathbf{x}||^2}$ is known as the conformal factor and $g^E$ is the Euclidean metric tensor defined as $g^E = \text{diag}([1,1,\ldots,1])$.
The distance between two points $\mathbf{x},\mathbf{y} \in \mathbb{D}^n$ is computed as $d_{\mathbb{D}}(\mathbf{x},\mathbf{y}) = \text{cosh}^{-1} \Big( 1 + 2 \frac{||\mathbf{x}-\mathbf{y}||^2}{(1-||\mathbf{x}||^2)(1-||\mathbf{y}||^2)} \Big)$.
For any point $\mathbf{x} \in \mathbb{D}^n$, let $\mathcal{T}_\mathbf{x} \mathbb{D}^n$ denote the associated tangent space, which is a subset of the Euclidean space.
We use the exponential map $\exp_x \colon \mathcal{T}_\mathbf{x} \mathbb{D}^n \rightarrow \mathbb{D}^n$ and the logarithmic map $\log_\mathbf{x} \colon \mathbb{D}^n \rightarrow \mathcal{T}_\mathbf{x} \mathbb{D}^n$ to map points from the tangent space to the hyperbolic space and from the hyperbolic space to the tangent space, respectively.
For the tangent vector $\mathbf{v} \neq \mathbf{0}$, we have $\exp_\mathbf{x}(\mathbf{v}) = \mathbf{x} \oplus \bigg( \text{tanh}\Big(\frac{\lambda_\mathbf{x} ||\mathbf{v}||}{2} \Big) \frac{\mathbf{v}}{||\mathbf{v}||} \bigg)$.
For the point $\mathbf{y} \neq \mathbf{0}$, we have $\log_\mathbf{x}(\mathbf{y}) = \frac{2}{\lambda_\mathbf{x}} \text{arctanh} \big(||-\mathbf{x} \oplus \mathbf{y}|| \big) \frac{-\mathbf{x} \oplus \mathbf{y}}{||-\mathbf{x} \oplus \mathbf{y}||}$
where $\oplus$ is the Mobius addition for any $\mathbf{x}, \mathbf{y} \in \mathbb{D}^n$ (hyperbolic analogous to vector addition in Euclidean space): $\mathbf{x} \oplus \mathbf{y} = \frac{(1 + 2 \langle \mathbf{x}, \mathbf{y} \rangle + ||\mathbf{y}||^2) \mathbf{x} + (1-||\mathbf{x}||^2) \mathbf{y}}{1 + 2 \langle \mathbf{x}, \mathbf{y} \rangle + ||\mathbf{x}||^2 ||\mathbf{y}||^2}$.
Similar to previous works, we choose the origin as the point on the manifold in whose tangent space we operate.

\section{Weisfeiler-Leman Hyperbolic Network}\label{sec:contribution}

\subsection{Embedding WL hierarchies}
As already discussed, prior GNNs largely ignore the distance between nodes.
It is not thus clear whether two nodes that have structurally dissimilar neighborhoods will obtain dissimilar representations.
Likewise, there are no guarantees that two nodes whose neighborhoods are structurally very similar will be mapped close to each other by the intermediate layers of a model.
However, capturing such distances between nodes is of paramount importance for machine learning applications since in node classification, similar nodes usually belong to the same class, while in node regression, similar nodes are associated with similar target values.
A natural question then is: how is the distance between two nodes defined?
Unfortunately, there is no clear answer to the above question.
Several different similarity and distance functions were proposed over the past decades for comparing graphs, subgraphs or nodes (\ie subgraphs centered at nodes), however, most of those functions are hard to compute.
For instance, the maximum common subgraph problem is known to be NP-hard~\citep{garey1979computers}.
Since non-polynomial time computable functions are of no practical use, we focus on a distance function of nodes which can be derived from the hierarchy generated by the WL algorithm. 
This distance function has already been studied in previous works~\citep{kriege2019computing,togninalli2019wasserstein}.
Formally, let $H=(V,E)$ be a rooted tree representing the hierarchy produced by the WL algorithm.
Let also $d_{\text{WL}} : V \times V \rightarrow \mathbb{R}_{\geq 0}$ denote the WL distance of nodes of tree $H$ defined as $d_{\text{WL}}(v,u) = \sum_{e \in P(v,u)} 1$ where $P(v, u)$ is the unique path from $v$ to $u$, for all $v,u \in V$. 
It is trivial to show that $d_{\text{WL}}$ is a metric on $V$.

Producing representations that preserve the distances between the nodes of the hierarchy is then equivalent to embedding the hierarchy into some space.
Unfortunately, it is not possible to embed trees into any Euclidean space with arbitrarily low distortion~\citep{linial1995geometry}.
This highlights a limitation of Euclidean GNNs which might fail to accurately encode into the learned node representations the information contained in the hierarchy produced by the WL algorithm.
On the other hand, it has been shown that trees can be embedded into the Poincar{\'e} disk $\mathbb{D}^2$ with arbitrarily low distortion~\citep{sarkar2011low}.
This motivates the use of models that embed nodes into the hyperbolic space and which can naturally encode the information contained in hierarchical structures.
Note, however, that most existing approaches for learning hyperbolic embeddings assume that there exists a latent hierarchy in the input data and their objective is to learn that hierarchy~\citep{liu2019hyperbolic,chami2019hyperbolic}.
In our setting, the hierarchy is known in advance, and the objective is to directly embed the hierarchy into some space.

\noindent\textbf{Sarkar's construction and limitations.}
There exists a construction proposed by \cite{sarkar2011low} which we can leverage to embed the nodes of the hierarchy into the Poincar{\'e} disk $\mathbb{D}^2$ and preserve the WL distance with arbitrarily low distortion.
However, Sarkar's construction (more details are given in Appendix~\ref{sec:sarkar}) is purely combinatorial and involves no learning (thus cannot be part of an end-to-end learning model).
Furthermore, the number of bits of precision used to represent components of the embedded nodes scales linearly with the maximum path length~\citep{sala2018representation}.
Thus, Sarkar's construction might experience numerical instabilities in the case of trees that contain long paths.
\cite{sala2018representation} generalized Sarkar's construction from the Poincare disk $\mathbb{D}^2$ to the Poincare ball $\mathbb{D}^d$ to deal with such problems, but still, the new construction is also combinatorial.
The above algorithms aim to maintain an equal distance between the children of a given node.
In other words, these algorithms equally space the children of a node around a circle.
However, in our setting, the hierarchy that is produced by the WL algorithm is not entirely unordered.
For instance, given three nodes $v,u,w$, if $\text{deg}(v) < \text{deg}(u) < \text{deg}(w)$, then the three nodes belong to different color classes and $v$ is more similar to $u$ than to $w$.
Such relationships between the different color classes can be captured by a MPNN model which iteratively updates the representations of the nodes.

\noindent\textbf{Proposed embedding approach.}
To take into account both the hierarchy produced by the WL algorithm and the aforementioned relationships between the nodes, and to also allow the embedding algorithm to be integrated into the pipeline of a learning model, we generalize Sarkar's algorithm.
The proposed construction which is shown in Algorithm~\ref{alg:construction} uses a node placement step that is based on the representations generated by a MPNN (line 4).
Specifically, we first normalize those representations such that they lie on the surface of the unit hypersphere (line 6).
Then, the emerging representations are scaled by $\tau$ and the children are placed in the corresponding positions in the hypersphere (lines 10 and 12).
In contrast to previous works~\citep{sarkar2011low, sala2018representation}, we do not space out children of a given node, but we allow them to be placed close to each other in the hyperbolic space.
Unfortunately, while angles are preserved between hyperbolic and Euclidean space, distances are not preserved.
Thus, the distances derived from the Euclidean representations produced by the MPNN are not preserved.
However, as we show next, relationships between distances are preserved. 
\begin{proposition}
    Let $v_1, v_2, v_3$ denote the children of some node in the WL hierarchy, and let $\textbf{v}_1, \textbf{v}_2, \textbf{v}_3 \in \mathbb{R}^n$ denote some transformation of their Euclidean representations such that $||\textbf{v}_1|| = || \textbf{v}_2|| = || \textbf{v}_3|| < 1$.
    Then, if $||\textbf{v}_1 - \textbf{v}_2|| \leq ||\textbf{v}_1 - \textbf{v}_3||$ holds, $d_\mathbb{D}(\textbf{v}_1, \textbf{v}_2) \leq d_\mathbb{D}(\textbf{v}_1, \textbf{v}_3)$ also holds.
    \label{prop:prop1}
\end{proposition}
Based on the above result, and since isometric reflections across geodesics preserve hyperbolic distances, the final hyperbolic representations of the children of a given node preserve their corresponding distance relations from the Euclidean space.
It is also necessary to separate the children of a given node from the node's parent reflected representation.
We can obtain an angle at least equal to $\theta = \frac{\pi}{2}$ between the parent of the node and each child as follows: we can ensure that the components of the Euclidean representations of the children contain non-negative values by applying the ReLU function.
We can then set the representation of the parent equal to $[-1/\sqrt{d},-1/\sqrt{d},\ldots,-1/\sqrt{d}]^\top \in \mathbb{R}^d$ and rotate the hypersphere so that this vector is located at node's parent actual representation (line 8).

\begin{algorithm}[t]
   \caption{Proposed Differentiable Hyperbolic Construction (DiffHypCon)}
   \label{alg:construction}
    \begin{algorithmic}[1]
       \STATE {\bfseries Input:} Euclidean representations of current iteration $\mathbf{H}^{(t)}$, hyperbolic representations of previous two iterations $\mathbf{Z}^{(t-1)}$, $\mathbf{Z}^{(t-2)}$, scaling factor $\tau$, vector $\mathbf{w} = [-1/\sqrt{d},\ldots,-1/\sqrt{d}]^\top$
       \FOR{\textbf{each} unique row $\mathbf{z}_a$ of $\mathbf{Z}^{(t-1)}$ with parent  $\mathbf{z}_b$ from $\mathbf{Z}^{(t-2)}$}
       \STATE $(\mathbf{0}, \mathbf{u}) \gets \text{reflect}_{\mathbf{z}_a \rightarrow \mathbf{0}}(\mathbf{z}_a, \mathbf{z}_b)$
       \STATE $\{\mathbf{h}_{c_1},\ldots, \mathbf{h}_{c_{\text{deg}(a)-1}}\} \gets \text{children}_a(\mathbf{H}^{(t)})$ \hfill \{ retrieve representations of children of $a$ \}
       \FOR{$i=1$ {\bfseries to} $\text{deg}(a)-1$}
       \STATE $\tilde{\mathbf{h}}_{c_i} \gets \mathbf{h}_{c_i}/||\mathbf{h}_{c_i}||$ \hfill \{ project to the surface of the unit sphere \}
       \ENDFOR
       \STATE $(\mathbf{u}, \mathbf{v}_{c_1}, \ldots, \mathbf{v}_{c_{\text{deg}(a)-1}}) \gets$ $\text{reflect\_through\_zero}_{\mathbf{w} \rightarrow \mathbf{u}}(\mathbf{w}, \tilde{\mathbf{h}}_{c_1}, \ldots, \tilde{\mathbf{h}}_{c_{\text{deg}(a)-1}})$
       \FOR{$i=1$ {\bfseries to} $\text{deg}(a)-1$}
       \STATE $\tilde{\mathbf{v}}_{c_i} \gets \tanh(\tau/2) \, \mathbf{v}_{c_i}$ \hfill \{ scale edges by a factor $\tau$ \}
       \ENDFOR
       \STATE $(\mathbf{z}_a, \mathbf{z}_b, \mathbf{z}_{c_1}, \ldots, \mathbf{z}_{c_{\text{deg}(a)-1}}) \gets$ $\text{reflect}_{\mathbf{0} \rightarrow \mathbf{z}_a}(\mathbf{0}, \mathbf{u}, \tilde{\mathbf{v}}_{c_1}, \ldots, \tilde{\mathbf{v}}_{c_{\text{deg}(a)-1}})$
       \STATE  $\mathbf{Z}^{(t)} \gets \{\mathbf{z}_{c_1}, \ldots, \mathbf{z}_{c_{\text{deg}(a)-1}}\}$ \hfill \{ store hyperbolic representations in corresponding rows of $\mathbf{Z}^{(t)}$ \}
       \ENDFOR
       \STATE {\bfseries Output:} Embedded vectors $\mathbf{Z}^{(t)}$ in $\mathbb{D}^n$
    \end{algorithmic}
\end{algorithm}

\begin{algorithm}[t]
   \caption{Proposed WLHN Model}
   \label{alg:wlhn}
    \begin{algorithmic}[1]
       \STATE {\bfseries Input:} Adjacency matrix $\mathbf{A}$ and matrix of node features $\mathbf{X}$ of graph $G$, number of iterations $T$, scaling factor $\tau$
       \STATE $\mathbf{Z}^{(-1)} \gets \mathbf{0}$ \hfill \{ root of the tree \}
       \STATE $\mathbf{H}^{(0)} \gets \mathbf{X}$
       \STATE $\mathbf{Z}^{(0)} \gets \textsc{DiffHypCon}(\mathbf{H}^{(0)}, \mathbf{Z}^{(-1)}, \mathbf{Z}^{(-1)}, \tau)$ \hfill \{ initial hyperbolic representations \}
       \FOR{$i=1$ {\bfseries to} $T$}
       \STATE $\mathbf{H}^{(i)} \gets \textsc{MPNN}(\mathbf{A}, \mathbf{H}^{(i-1)})$ \hfill \{ use MPNN to update node representations \}
       \STATE $\mathbf{Z}^{(i)} \gets \textsc{DiffHypCon}(\mathbf{H}^{(i)}, \mathbf{Z}^{(i-1)}, \mathbf{Z}^{(i-2)}, \tau)$ \hfill \{ compute hyperbolic representations \}
       \ENDFOR
       \STATE {\bfseries Output:} Embeddings of nodes $\mathbf{Z}^{(T)}$ in $\mathbb{D}^n$
    \end{algorithmic}
\end{algorithm}

\subsection{Model Architecture}

\noindent\textbf{Neighborhood aggregation.}
We assume that the initial feature vectors of the nodes are from a countable universe.
For finite graphs, node representations at deeper layers of a MPNN are also from a countable universe~\citep{xu2019powerful}.
Thus, each node representation can be mapped to a color.
Then, the node representations that emerge at the different iterations of the message passing procedure correspond to different colors and from those colors we can construct a hierarchy.
In this paper, we use the neighborhood aggregation scheme of the GIN model~\citep{xu2019powerful} which is defined as follows:
\begin{equation*}
    \mathbf{h}_v^{(t)} = \text{MLP}^{(t)} \bigg( \big( 1+\epsilon^{(t)} \big) \, \mathbf{h}_v^{(t-1)} + \sum_{u \in \mathcal{N}(v)} \mathbf{h}_u^{(t-1)}  \bigg)
\end{equation*}
For each depth $t \in [T]$ of the hierarchy, each node belongs to one of the color classes of that depth.
Two nodes $v,u$ belong to the same color class of depth $t$ if $\mathbf{h}_v^{(t)} = \mathbf{h}_u^{(t)}$ holds.
Note that the GIN model can be as powerful as the WL algorithm in terms of distinguishing non-isomorphic graphs, and thus the emerging hierarchy could potentially be identical to the one produced by the WL algorithm (\ie in case proper weights are found).

Note that we do not construct a single hierarchy consisting of the color classes of all the nodes of all graphs.
Instead, a different hierarchy is constructed and embedded for each batch of samples.
This does not pose any problem since the proposed construction does not space out children and hence, it does not make any assumptions about the number of children of a given node. 
Let $\mathcal{H}_\mathcal{B}^{(t)} = \{ \mathbf{h}_v^{(t)} \mid v \in V_\mathcal{B} \}$ denote the set that contains the representations of all the nodes of the graphs contained in a batch $\mathcal{B}$ after the $t$-th iteration of the neighborhood aggregation procedure where $V_\mathcal{B}$ is the set of nodes of the graphs contained in $\mathcal{B}$.
These representations form the color classes of the hierarchy produced by the WL algorithm after its $t$-th iteration.
Specifically, there is a one-to-one mapping between those vectors and the nodes of the hierarchy whose path distance from the root node is equal to $t$ (or $t+1$ in the case of non-monochromatic initial coloring).
Furthermore, $|\mathcal{H}^{(t)}| \leq |\mathcal{H}^{(t+1)}|$ holds, and each representation has a parent.
For instance, the parent of $\mathbf{h}_v^{(t)}$ is $\mathbf{h}_v^{(t-1)}$.
Therefore, we can use the construction presented above to embed those representations into the hyperbolic space: $\mathbf{z}_v^{(t)} = \text{DiffHypCon}(\mathbf{h}_v^{(t)}, \mathbf{z}_v^{(t-1)}, \mathbf{z}_v^{(t-2)})$ where $\mathbf{z}_v^{(t)}$ is the hyperbolic representation of node $v$ in the $t$-th iteration of the neighborhood aggregation procedure.
Thus, each neighborhood aggregation operation is followed by an embedding phase where the emerging node representations are mapped to the Poincar{\'e} ball using the proposed construction.
The proposed method is illustrated in Algorithm~\ref{alg:wlhn}.

\noindent\textbf{Readout.}
To perform graph classification, we map the output of the last neighborhood aggregation layer of the model to the tangent space using the logarithmic map and aggregate the node representations using the sum operator.
\begin{equation*}
    \mathbf{h}_G = \sum_{v \in V} \log_\mathbf{0} \big( \mathbf{z}_v^T \big)
\end{equation*}
Then, the graph representation $\mathbf{h}_G$ is fed to further layers (\eg Euclidean multinomial logistic regression) to produce the output.
Alternatively, the rest of the computations can also be realized in the hyperbolic space, \ie we can directly classify samples on the hyperboloid manifold using the hyperbolic multinomial logistic loss~\citep{ganea2018hyperbolic}. 
In this case, we need to map the tangent vector back to the hyperbolic space using the exponential map $\mathbf{z}_G = \exp_\mathbf{0} \big( \mathbf{h}_G \big)$.
Thus, this model first maps node representations to the tangent space, it performs the node aggregation in the tangent space, and then maps the graph representation back to the hyperbolic space.
In preliminary experiments, we observed that this approach performs similarly to Euclidean classification.

Note that if the generated node representations accurately capture the distances of nodes, then the emerging graph representations also capture the distances of graphs.
\begin{proposition}
    Let $G_1=(V_1, E_1)$, $G_2=(V_2,E_2)$ denote two graphs.
    Without loss of generality, we assume that the two graphs have the same number of nodes, \ie $|V_1|=|V_2|=n$.
    Let $\mathbf{v}_i, \mathbf{u}_j$, $i,j \in [n]$ denote the representations of the nodes of the two graphs.
    Let also $\mathbf{h}_{G_1}$ and $\mathbf{h}_{G_2}$ denote the vector representations of $G_1$ and $G_2$, respectively, which emerge by applying the sum operator to the representations of the nodes of the two graphs.
    Then, we have that:
    \begin{equation*}
        || \mathbf{h}_{G_1} - \mathbf{h}_{G_2} || <= \min_\mathbf{T} \sum_{i=1}^n \sum_{j=1}^n \mathbf{T}_{ij} || \mathbf{v}_i - \mathbf{u}_j ||
    \end{equation*}
    where $\mathbf{T} \in \{ 0, 1 \}^{n \times n}$, $\mathbf{T} \, \mathbf{1} = \mathbf{1}$ and $\mathbf{1} \, \mathbf{T} = \mathbf{1}$.
    \label{prop:prop2}
\end{proposition}
The above Proposition implies that given two sets of vectors, the difference of the output of the sum aggregator for the two sets can be bounded via the stable matching between the two sets. 
Thus, if two graphs consist of similar nodes (in terms of their representations), the graph representations will also be similar to each other.
 
\noindent\textbf{Model depth.}
It has been reported that MPNNs usually do not benefit from more than few neighborhood aggregation layers.
This is attributed to different phenomena such as over-smoothing~\citep{li2018deeper}.
We should note that in contrast to Euclidean MPNNs, a large number of neighborhood aggregation layers does not have a negative impact on the performance of the proposed model since the learned node representations respect the structure of the WL hierarchy, and thus distances between nodes are preserved no matter how large the depth of the hierarchy is.
Furthermore, no residual connections are required since the final node representations encode the entire structure of the WL hierarchy, \ie the whole history of previous node representations.

\noindent\textbf{Computational complexity.}
The proposed model uses the neighborhood aggregation mechanism of GIN to update the representations of the nodes and then uses the DiffHypCon algorithm to embed the nodes into the hyperbolic space.
DiffHypCon places points into the unit hypersphere in linear time.
Therefore, the computational complexity of the model is comparable with that of other standard MPNNs and is in the order of $\mathcal{O}(Tmd^2)$ where $d$ denotes the dimension of the node features.
We empirically measured the running time of the proposed model on real-world datasets and we report the results in the Appendix.

\section{Experimental Evaluation}\label{sec:experiments}

\subsection{Synthetic Datasets}
\begin{figure}[t]
    \centering
    \includegraphics[width=.7\linewidth]{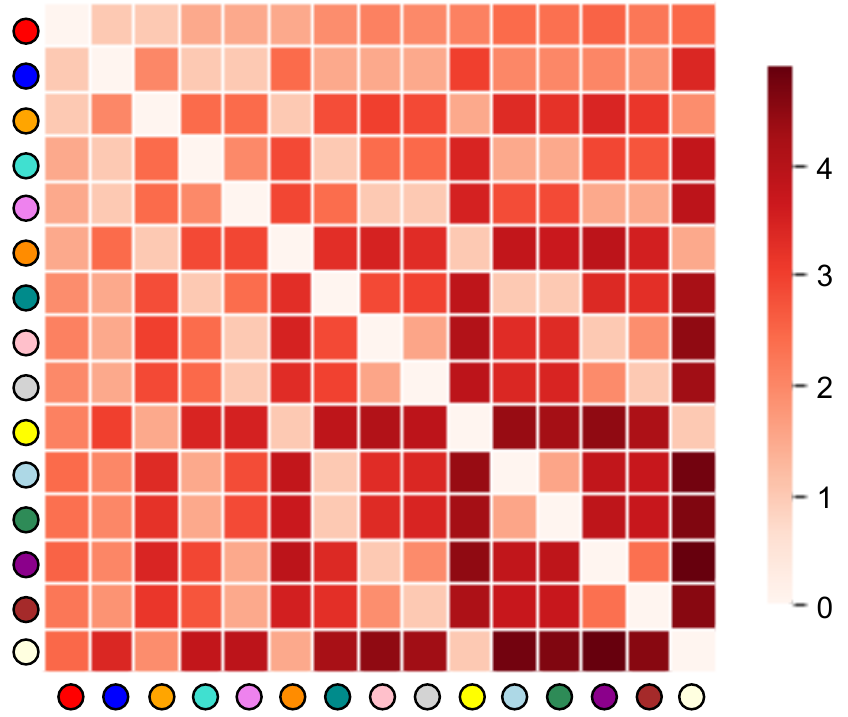}
    \caption{Heatmap that illustrates the distances between all pairs of nodes of the hierarchy of Figure~\ref{fig:example}.
    Distances were computed between the nodes' generated hyperbolic representations.}
    \label{fig:heatmap}
\end{figure}

\noindent\textbf{Qualitative results.}
To empirically verify that the proposed model can encode the hierarchy $H_G$ created by the WL algorithm, we applied it to the graph shown in Figure~\ref{fig:example}.
Specifically, we randomly initialized the parameters of the model and we performed a feedforward pass (no training was performed) to generate the graph representations that lie in the hyperbolic space ($\tau$ was set equal to $1$).
Then, given the hyperbolic representations of the nodes of the hierarchy, we computed the distance between the different nodes in the hyperbolic space.
The emerging distances were then compared against the WL distance of nodes $d_{\text{WL}}$ defined in section~\ref{sec:contribution}.
We computed the Pearson correlation coefficient which was found to be equal to $0.98$.
We also retrieved the Euclidean representations of the nodes (produced by the GIN model).
We computed the Euclidean distance between those representations and we also compared the emerging distances against the WL distance.
In this case, the correlation turned out to be much smaller and equal to $0.35$.
Hence, the emerging hyperbolic representations lead to distances that are much more correlated with the distances derived from the WL hierarchy compared to those that emerge from the Euclidean representations.
We also produced a heatmap (shown in Figure~\ref{fig:heatmap}) that illustrates the distances between the nodes of the hierarchy which were computed based on the nodes' hyperbolic representations.
We can see that the obtained distances are similar to the corresponding WL distances, while the structure is preserved since the nodes closest to a given node are its direct neighbors in the hierarchy.
For example, the blue and the red nodes are neighbors in the hierarchy $H_G$ while the distance of their hyperbolic representations is small. 
 
\begin{table}[t]
    \centering
    \scriptsize
    \def\arraystretch{1.1}
    \caption{MSE ± standard deviation results of the proposed model and the baselines in the task of predicting the effective size and the density of nodes in different types of graphs.}
    \label{tab:structural_properties}
    \begin{tabular}{lccc}
    \toprule
    \multirow{2}{*}{} & \multicolumn{2}{c}{\textbf{Density (Barab{\'a}si-Albert graphs)}} \\
    & m=5 & m=10 \\ 
    \midrule
    GCN & 0.0193 $\pm$ 0.0026 & 0.0177 $\pm$ 0.0023 \\
    GIN & 0.0113 $\pm$ 0.0032 & 0.0162 $\pm$ 0.0031 \\
    \hline
    \textbf{WLHN} & \textbf{0.0034} $\pm$ 0.0004 & \textbf{0.0024} $\pm$ 0.0003 \\
    \bottomrule
    \end{tabular}\\
    \vspace{.1cm}
    \begin{tabular}{lccc}
    \toprule
    \multirow{2}{*}{} & \multicolumn{2}{c}{\textbf{Effective Size (Erd{\"o}s-R{\'e}nyi graphs)}} \\
    & p=0.008 & p=0.01  \\
    \midrule
    GCN & 0.0088 $\pm$ 0.0026 & 0.0087 $\pm$ 0.0028 \\
    GIN & 0.0017 $\pm$  0.0004 & 0.0015 $\pm$ 0.0006 \\
    \hline
    \textbf{WLHN} & \textbf{0.0006} $\pm$ 0.0003 & \textbf{0.0014} $\pm$ 0.0008 \\
    \bottomrule
    \end{tabular}
 \end{table}

\noindent\textbf{Quantitative results.}
We also constructed two node regression datasets where the task is to predict the density of the nodes' ego-networks and their effective size, respectively.
The first dataset contains $10$ Barab{\'a}si-Albert graphs.
The number of nodes of each graph is set to $1,000$ and  the number of edges to attach from a new node to existing nodes is set to $5$ or $10$ (\ie $m=5$ or $m=10$).
We compute the target of each node as follows: we extract the nodes' ego-networks of radius $2$ and compute their densities (the density of a graph consisting of $n$ nodes and $m$ edges is equal to $m/n$).
The second dataset contains $10$ Erd{\"o}s-R{\'e}nyi graphs.
The number of nodes is set to $1,000$ and the probability for edge creation is set to either $0.008$ or $0.01$ (\ie $p=0.008$ or $p=0.01$).
The target of each node is set equal to its effective size which for a node $v$ is defined as $e(v) = \text{deg}(v) - 2t/\text{deg}(v)$ where $t$ is the number of edges between neighbors of $v$.
We split the nodes of each graph into training/validation/test sets with a ratio of $60\%/20\%/20\%$.
We trained the models by minimizing the mean squared error (MSE) loss.
We repeated the whole experiment $10$ times and we report in Table~\ref{tab:structural_properties} the average MSE and corresponding standard deviation.
We can see that the proposed model outperforms the two baselines (GCN~\citep{kipf2017semi} and GIN~\citep{xu2019powerful}) under all settings.
The two predicted properties (\ie density and effective size) are related to the structure of the neighborhood of each node, and thus the experimental results verify our claim that WLHN can better capture the structural distance of nodes.

\begin{table*}[t]
    \centering
    \scriptsize
    \def\arraystretch{1}
    \caption{Classification accuracy $\pm$ standard deviation of the proposed model and the baselines on the $5$ node classification datasets.}
    \label{tab:node_classification_results}
    \begin{tabular}{lccccc}
    \toprule & \textbf{Cornell} & \textbf{Texas} & \textbf{Wisconsin} & \textbf{Squirrel} & \textbf{Actor}    \\ 
    \midrule          
    GCN & 60.54 $\pm$ 5.30 & 55.14 $\pm$ 5.16 & 51.76 $\pm$ 3.06 & 53.43 $\pm$ 2.01 & 27.32 $\pm$ 1.10 \\   
    GIN & 51.62 $\pm$ 7.68 & 53.24 $\pm$ 8.55 & 50.59 $\pm$ 7.98 & 44.12 $\pm$ 2.04 & 29.67 $\pm$ 0.64\\ 
    GAT & 61.89 $\pm$ 5.05 & 52.16 $\pm$ 6.63 & 49.41 $\pm$ 4.09 & 40.72 $\pm$ 1.55 & 27.44 $\pm$ 0.89  \\
    MixHop & 73.51 $\pm$ 6.34 & \textbf{77.84} $\pm$ 7.73 & 75.88 $\pm$ 4.90 & 43.80 $\pm$ 1.48 & 32.22 $\pm$ 2.34 \\
    HGCN (PoincareBall) & 53.51 $\pm$ 6.02 & 54.86 $\pm$ 6.95 & 63.53 $\pm$ 9.25 & 50.78 $\pm$ 1.52 & 31.66 $\pm$ 1.30\\
    HGCN (Hyperboloid) & 55.68 $\pm$ 7.76 & 53.24 $\pm$ 8.80 & 70.20 $\pm$ 7.73 & 44.87 $\pm$ 1.89 & 30.75 $\pm$ 4.33 \\
    Geom-GCN & 60.54 $\pm$ 3.67 & 66.76 $\pm$ 2.72 & 64.51$\pm$ 3.66 & 38.15 $\pm$ 0.92 & 34.59\\
    P-GNN & 74.32 $\pm$ 4.87 & 81.62 $\pm$ 6.60 & \textbf{83.52} $\pm$ 5.63 & 33.54 $\pm$  1.71 & 34.09 $\pm$ 1.00\\
    \midrule
    \textbf{WLHN} & \textbf{77.29} $\pm$ 4.66 & 75.41 $\pm$ 5.98 & 78.62 $\pm$ 3.44 & \textbf{55.76} $\pm$ 0.92 & \textbf{36.42} $\pm$ 1.42\\ 
    \bottomrule
    \end{tabular}   
\end{table*}

\begin{table*}[t]
    \caption{Classification accuracy $\pm$ standard deviation of the proposed model and the baselines on the $10$ benchmark datasets. OOR means Out of Resources, either time ($>$72 hours for a single training) or GPU memory.}
    \label{tab:classification_results}
    \centering
    \scriptsize
    \def\arraystretch{1.2}
    \addtolength{\tabcolsep}{-2.5pt}    
    \begin{tabular}{lcccccccccc}
    \toprule
    & \textbf{MUTAG} & \textbf{D\&D} & \textbf{NCI1} & \textbf{PROTEINS} & \textbf{ENZYMES} & \textbf{IMDB-B} & \textbf{IMDB-M} & \textbf{REDDIT-B} & \textbf{REDDIT-5K} & \textbf{COLLAB}  \\
    \midrule
    DGCNN & 84.0 $\pm$ 6.7 & 76.6 $\pm$ 4.3 & 76.4 $\pm$ 1.7 & 72.9 $\pm$ 3.5 & 38.9 $\pm$ 5.7 & 69.2 $\pm$ 3.0 & 45.6 $\pm$ 3.4 & 87.8 $\pm$ 2.5 & 49.2 $\pm$ 1.2 & 71.2 $\pm$ 1.9 \\ 
    DiffPool & 79.8 $\pm$ 7.1 & 75.0 $\pm$ 3.5 & 76.9 $\pm$ 1.9 & 73.7 $\pm$ 3.5 & 59.5 $\pm$ 5.6 & 68.4 $\pm$ 3.3 & 45.6 $\pm$ 3.4 & 89.1 $\pm$ 1.6 & 53.8 $\pm$ 1.4 & 68.9 $\pm$ 2.0 \\ 
    ECC & 75.4 $\pm$ 6.2 & 72.6 $\pm$ 4.1 & 76.2 $\pm$ 1.4 & 72.3 $\pm$ 3.4 & 29.5 $\pm$ 8.2 & 67.7 $\pm$ 2.8 & 43.5 $\pm$ 3.1 & OOR & OOR & OOR  \\ 
    GraphSAGE & 83.6 $\pm$ 9.6 & 72.9 $\pm$ 2.0 & 76.0 $\pm$ 1.8 & 73.0 $\pm$ 4.5 & 58.2 $\pm$ 6.0 & 68.8 $\pm$ 4.5 & 47.6 $\pm$ 3.5 & 84.3 $\pm$ 1.9 & 50.0 $\pm$ 1.3 & 73.9 $\pm$ 1.7 \\
    GCN & 79.2 $\pm$ 9.4 & 76.6 $\pm$ 4.0 & 76.8 $\pm$ 1.6 & 73.7 $\pm$ 2.9 & 57.2 $\pm$ 5.8 & 70.7 $\pm$ 5.3 &  47.9 $\pm$ 4.2 & 90.1 $\pm$ 2.1 & 55.0 $\pm$ 1.7 & 71.3 $\pm$ 2.0 \\
    GIN & 84.7 $\pm$ 6.7 & 75.3 $\pm$ 2.9 & \textbf{80.0} $\pm$ 1.4 & 73.3 $\pm$ 4.0 & 59.6 $\pm$ 4.5 & 71.2 $\pm$ 3.9 & 48.5 $\pm$ 3.3 & 89.9 $\pm$ 1.9 & \textbf{56.1} $\pm$ 1.7 & 75.6 $\pm$ 2.3 \\
    HGCN (PoincareBall) & 83.4 $\pm$ 6.7 & 78.0 $\pm$ 2.8 & 74.2 $\pm$ 2.4 & 74.4 $\pm$ 3.1 & 39.7 $\pm$ 5.5 & 73.0 $\pm$ 3.2 & \textbf{50.3} $\pm$ 3.8 & 87.9 $\pm$ 2.8 & 49.4 $\pm$ 2.6 & 80.2 $\pm$ 1.9  \\
    HGCN (Hyperboloid) & 83.4 $\pm 6.2$ & 77.8 $\pm 4.3$ & 72.3 $\pm$ 4.3 & 74.7 $\pm$ 3.4 & 32.3 $\pm$ 5.4 & 73.3 $\pm$ 3.5 & \textbf{50.3} $\pm$ 4.0 & 86.3 $\pm$ 1.6 & 52.7 $\pm$ 2.0 & \textbf{80.3} $\pm$ 1.8\\
    \midrule
    \textbf{WLHN} & \textbf{86.0} $\pm$ 7.4 & \textbf{78.5} $\pm$ 3.4 & 79.2 $\pm$ 1.1 & \textbf{75.9} $\pm$ 1.9 & \textbf{62.5} $\pm$ 5.0 & \textbf{73.4} $\pm$ 3.7 & 49.7 $\pm$ 3.6 & \textbf{90.7} $\pm$ 1.9 & 55.2 $\pm$ 1.2 & 76.2 $\pm$ 2.3\\
    \bottomrule
    \end{tabular}
    \addtolength{\tabcolsep}{2.5pt}    
\end{table*}

\begin{table}[t]
    \centering
    \scriptsize
    \caption{Performance of the proposed model and the baselines on the ogbg-molhiv and ogbg-molpcba datasets.}
    \label{tab:ogbg_results}
    \def\arraystretch{1.1}
    \begin{tabular}{lcc}
    \toprule
              & \textbf{ogbg-molhiv}     & \textbf{ogbg-molpcba}     \\ 
              & ROC-AUC & Avg. Precision \\
    \midrule          
    GCN       & 76.06 $\pm$ 0.97 & 20.20 $\pm$ 0.24 \\ 
    GIN       & 75.58 $\pm$ 1.40 & 22.66 $\pm$ 0.28 \\ 
    HGCN (PoincareBall) & 76.42 $\pm$ 1.75 & 17.73 $\pm$ 0.22\\
    HGCN (Hyperboloid) & 75.91 $\pm$ 1.48 & 17.52 $\pm$ 0.20 \\
    \midrule
    \textbf{WLHN} & \textbf{78.41} $\pm$ 0.31 & \textbf{22.90} $\pm$ 0.25 \\ 
    \bottomrule
    \end{tabular}
\end{table}
 
\subsection{Real-World Datasets}

\noindent\textbf{Datasets.}
We evaluated the proposed model on five node classification datasets: Cornell, Texas, Wisconsin, Squirrel, and Actor.
The first three datasets are extracted from the WebKB dataset and the rest of the datasets from Wikipedia, and have been employed in previous studies~\citep{pei2020geom,rozemberczki2021multi}.
We also evaluated the proposed model on standard graph classification datasets derived from the TUD repository~\citep{morris2020tudataset}.
We employed $5$ datasets from bioinformatics and chemoinformatics (MUTAG, D\&D, NCI1, PROTEINS, ENZYMES), and $5$ datasets from social networks (IMDB-BINARY, IMDB-MULTI, REDDIT-BINARY, REDDIT-MULTI-$5$K, COLLAB).
For datasets that contain graphs whose nodes are annotated with discrete labels, we map those labels to one-hot vectors.
We also evaluated the proposed model on two graph classification datasets from the Open Graph Benchmark (OGB)~\citep{hu2020open}, a collection of large-scale datasets.
Specifically, we used the following two molecular property prediction datasets: ogbg-molhiv and ogbg-molpcba.

\noindent\textbf{Baselines.}
In the node classification task, we compare WLHN against the following models: GAT~\citep{velikovi2017graph}, GCN~\citep{kipf2017semi}, GIN~\citep{xu2019powerful}, MixHop~\citep{abu2019mixhop} and HGCN~\citep{chami2019hyperbolic}.
In the graph classification task, we compare the proposed model against the following seven GNNs: DGCNN~\citep{zhang2018end}, DiffPool~\citep{ying2018hierarchical}, ECC~\citep{simonovsky2017dynamic}, GraphSAGE~\citep{hamilton2017inductive}, GCN~\citep{kipf2017semi}, GIN~\citep{xu2019powerful} 
and HGCN~\citep{chami2019hyperbolic}.
In the case of the OGB datasets, we compare the WLHN model against the last three of the above models.

\noindent\textbf{Experimental setup.}
For the node classification datasets, we randomly split each dataset into training/validation/test sets with a ratio of $60\%/20\%/20\%$, while the whole process was repeated $10$ times.
For the standard graph classification datasets, we perform $10$-fold cross-validation, where within each fold a model's hyperparameters are selected based on a $90\%/10\%$ split of the training set.
We use the evaluation framework from~\cite{errica2020fair}, thus employing the same split for each evaluated method.
For the two datasets from OGB, we use the standard splits associated with those datasets.
A detailed description of the hyperparameter selection approach is given in Appendix~\ref{sec:hyperparams}.

\noindent\textbf{Results.}
Table~\ref{tab:node_classification_results} presents the performance of the different models on the five node classification datasets.
We observe that WLHN is the best performing method since it outperforms the baselines on $3$ out of the $5$ datasets.
On these $3$ datasets, the proposed model improves significantly over the baseline models.
Specifically, the proposed model yields respective absolute improvements of $2.97\%$, $2.33\%$ and $1.83\%$ in accuracy over the best competitor on the Cornell, Squirrel and Actor datasets, respectively.
Table~\ref{tab:classification_results} illustrates the average accuracies and corresponding standard deviations for the TUD datasets.
We observe that the proposed model outperforms the baselines on $6$ out of the $10$ datasets.
On MUTAG, PROTEINS and ENZYMES, the proposed model provides significant performance gains compared to the baseline models.
The proposed model embeds the WL tree hierarchy produced by the GIN model into the hyperbolic space to produce expressive node representations.
On the other hand, our two main competitors, GIN and HGCN, have some serious limitations.
GIN operates in the Euclidean space where hierarchical structures cannot be accurately captured.
HGCN embeds nodes in the hyperbolic space but it does not have explicit access to the WL hierarchy, thus leading to less informed representations.
Finally, Table~\ref{tab:ogbg_results} illustrates the performance of the proposed model on the two OGB datasets.
On ogbg-molhiv, the proposed model outperforms all the baselines, most of them by wide margins.
On the other hand, on ogbg-molpcba, while the proposed model is still the best-performing method, some of the baselines achieve an average precision close to that of WLHN.
Overall, the results indicate that the proposed model exhibits competitive performance in both node and graph classification tasks.

\noindent\textbf{Additional Experiments.}
We further investigate the robustness of WLHN under structural and feature noise.
We observe that noise does not have a very large impact on the model's performance, while WLHN also significantly outperforms the baselines.
We also experimentally demonstrate that WLHN does not suffer from over-smoothing even if the number of neighborhood aggregation layers is large, since the proposed hyperbolic embedding algorithm preserves the distances of the nodes.
We present the results in the Appendix.

\section{Conclusion}\label{sec:conclusion}
In this paper, we proposed a new model which preserves distances of nodes/graphs, a problem that has not been investigated thoroughly so far.
Our model learns node representations in the hyperbolic space which respect the hierarchy generated by MPNNs.
We defined a distance function and proposed a novel GNN model which can accurately capture that distances by embedding the nodes of the graphs in the hyperbolic space.
The emerging distances also take into account the Euclidean representations of the nodes produced by the neighborhood aggregation procedure of a MPNN model.
We evaluated the proposed model on synthetic and real-world datasets.
Our results demonstrate that the proposed model can indeed encode meaningful distances in the learned representations, while it achieves high levels of performance in node and graph classification tasks.

\subsubsection*{Acknowledgements}
G.N. is supported by the French National research agency via the AML-HELAS (ANR-19-CHIA-0020) project.
The authors would like to thank the anonymous AISTATS reviewers for the constructive comments.

\bibliography{aistats2023}

\appendix
\onecolumn

The Appendix is organized as follows. 
In sections~\ref{sec:prop1} and~\ref{sec:prop2}, we prove Proposition~\ref{prop:prop1} and Proposition~\ref{prop:prop2}, respectively.
In section~\ref{sec:sarkar}, we give more details about Sarkar's construction.
In sections~\ref{sec:add_experiments} and~\ref{sec:hyperparams}, we present a set of additional experiments we performed (measuring running time, robustness to noise and to the problem of oversmoothing) and describe the hyperparameters we used in our experiments, respectively.
In section~\ref{sec:datasets}, we provide statistics and descriptions of the datasets we used in our experiments.

\section{Proof of Proposition~\ref{prop:prop1}}\label{sec:prop1}
We have that $||\textbf{v}_1|| = ||\textbf{v}_2|| = ||\textbf{v}_3||$ and also that $||\textbf{v}_1 - \textbf{v}_2|| \leq ||\textbf{v}_1 - \textbf{v}_3||$.
Then, we have:
\begin{equation*}
    \begin{split}
        ||\textbf{v}_1 - \textbf{v}_2|| \leq ||\textbf{v}_1 - \textbf{v}_3|| &\implies ||\textbf{v}_1 - \textbf{v}_2||^2 \leq ||\textbf{v}_1 - \textbf{v}_3||^2 \\
        &\implies \frac{||\mathbf{v}_1-\mathbf{v}_2||^2}{(1-||\mathbf{v_1}||^2)(1-||\mathbf{v}_2||^2)} \leq \frac{||\mathbf{v}_1-\mathbf{v}_3||^2}{(1-||\mathbf{v_1}||^2)(1-||\mathbf{v}_3||^2)} \\
        &\implies 1 + 2\frac{||\mathbf{v}_1-\mathbf{v}_2||^2}{(1-||\mathbf{v_1}||^2)(1-||\mathbf{v}_2||^2)} \leq 1 + 2\frac{||\mathbf{v}_1-\mathbf{v}_3||^2}{(1-||\mathbf{v_1}||^2)(1-||\mathbf{v}_3||^2)} \\
        &\implies \text{cosh}^{-1} \Big(1 + 2\frac{||\mathbf{v}_1-\mathbf{v}_2||^2}{(1-||\mathbf{v_1}||^2)(1-||\mathbf{v}_2||^2)} \Big) \leq \text{cosh}^{-1} \Big( 1 + 2\frac{||\mathbf{v}_1-\mathbf{v}_3||^2}{(1-||\mathbf{v_1}||^2)(1-||\mathbf{v}_3||^2)} \Big) \\
        &\implies d_{\mathbb{D}}(\mathbf{v}_1,\mathbf{v}_2) \leq d_{\mathbb{D}}(\mathbf{v}_1,\mathbf{v}_3)
    \end{split}
\end{equation*}
which concludes the proof.

\section{Proof of Proposition~\ref{prop:prop2}}\label{sec:prop2}
Let $G_1=(V_1, E_1)$ and $G_2=(V_2,E_2)$ denote two graphs.
Without loss of generality, we assume that the two graphs have the same number of nodes, \ie $|V_1|=|V_2|=n$.
Let $\mathcal{H}_1$ and $\mathcal{H}_2$ denote the set of vector representations of the nodes of $G_1$ and $G_2$, respectively.
Then, $\mathcal{H}_1$ and $\mathcal{H}_2$ have the same cardinality, thus $|\mathcal{H}_1|=|\mathcal{H}_2|$ holds.
Let also $\mathcal{H}_1 = \{ \mathbf{v}_1,\ldots, \mathbf{v}_n\}$ and $\mathcal{H}_2 = \{ \mathbf{u}_1,\ldots, \mathbf{u}_n\}$.
Then, we have:
\begingroup
\allowdisplaybreaks
\begin{align*}
       || \mathbf{h}_{G_1} - \mathbf{h}_{G_2} || &= \bigg|\bigg| \sum_{i=1}^n \mathbf{v}_i -  \sum_{i=1}^n \mathbf{u}_i \bigg|\bigg| \\
        &=|| \mathbf{v}_1+\mathbf{v}_2+\ldots+\mathbf{v}_n-\mathbf{u}_1-\mathbf{u}_2-\ldots-\mathbf{u}_n|| \\
        &= || \mathbf{v}_1-\mathbf{u}_{f(v_1)} + \mathbf{v}_2 - \mathbf{u}_{f(v_2)} + \ldots + \mathbf{v}_n - \mathbf{u}_{f(v_n)} || \\
        &\leq || \mathbf{v}_1-\mathbf{u}_{f(v_1)}|| + ||\mathbf{v}_2 - \mathbf{u}_{f(v_2)}|| + \ldots + ||\mathbf{v}_n - \mathbf{u}_{f(v_n)} || \\
        &= \min_\mathbf{T} \sum_{i=1}^n \sum_{j=1}^n \mathbf{T}_{ij} || \mathbf{v}_i - \mathbf{u}_j || \\
        &\qquad \qquad \quad \text{s.t.}\\
        &\qquad \qquad \mathbf{T} \in \{ 0, 1 \}^{n \times n}\\
        &\qquad \qquad \mathbf{T} \, \mathbf{1} = \mathbf{1}\\
        &\qquad \qquad \mathbf{1} \, \mathbf{T} = \mathbf{1} 
\end{align*}
\endgroup
which concludes the proof.
Note that the right part of the last equality above is the solution of the following bipartite matching problem.
We construct a complete bipartite graph where the first and second partition contain nodes of $G_1$ and $G_2$, respectively, while the weight of an edge is some positive value inversely proportional to the Euclidean distance of the vector representations of the two endpoints.
Then, matrix $\mathbf{T}$ is the solution of the maximum bipartite matching on the above graph, thus it can be seen as a function that maps nodes of $G_1$ to matched nodes of $G_2$.

\begin{figure}[t]
\begin{minipage}{.5\textwidth}
    \begin{algorithm}[H]
    \hsize=\textwidth
  \caption{Sarkar's Construction}
  \label{alg:sarkar}
  \begin{algorithmic}[1]
      \STATE {\bfseries Input:} Node $a$ with parent $b$, children to place $c_1, c_2, \ldots, c_{\text{deg}(a)-1}$, partial embedding $f$ containing an embedding for $a$ and $b$, scaling factor $\tau$
      \STATE $(0, z) \gets \text{reflect}_{f(a) \rightarrow 0}(f(a), f(b))$
      \STATE $\theta \gets \text{arg}(z)$ \hfill \{ angle of $z$ from $x$-axis in the plane \}
      \FOR{$i=1$ {\bfseries to} $\text{deg}(a)-1$}
      \STATE $y_i \gets \frac{\exp^\tau+1}{\exp^\tau-1} \Big( \cos \big(\theta+\frac{2\pi i}{\text{deg}(a)} \big), \sin \big( \theta + \frac{2\pi i}{\text{deg}(a)} \big) \Big)$
      \ENDFOR
      \STATE $(f(a), f(b), f(c_1), f(c_2), \ldots,f(c_{\text{deg}(a)-1})) \gets$ $\text{reflect}_{0 \rightarrow f(a)}(0, z, y_1, y_2, \ldots, y_{\text{deg}(a)-1})$
      \STATE {\bfseries Output:} Embedded $\mathbb{D}^2$ vectors $f(c_1), f(c_2), \ldots,$ $f(c_{\text{deg}(a)-1})$
    \end{algorithmic}
\end{algorithm}
\end{minipage}\hfill
\begin{minipage}{.4\textwidth}
  \centering
    \includegraphics[width=\linewidth]{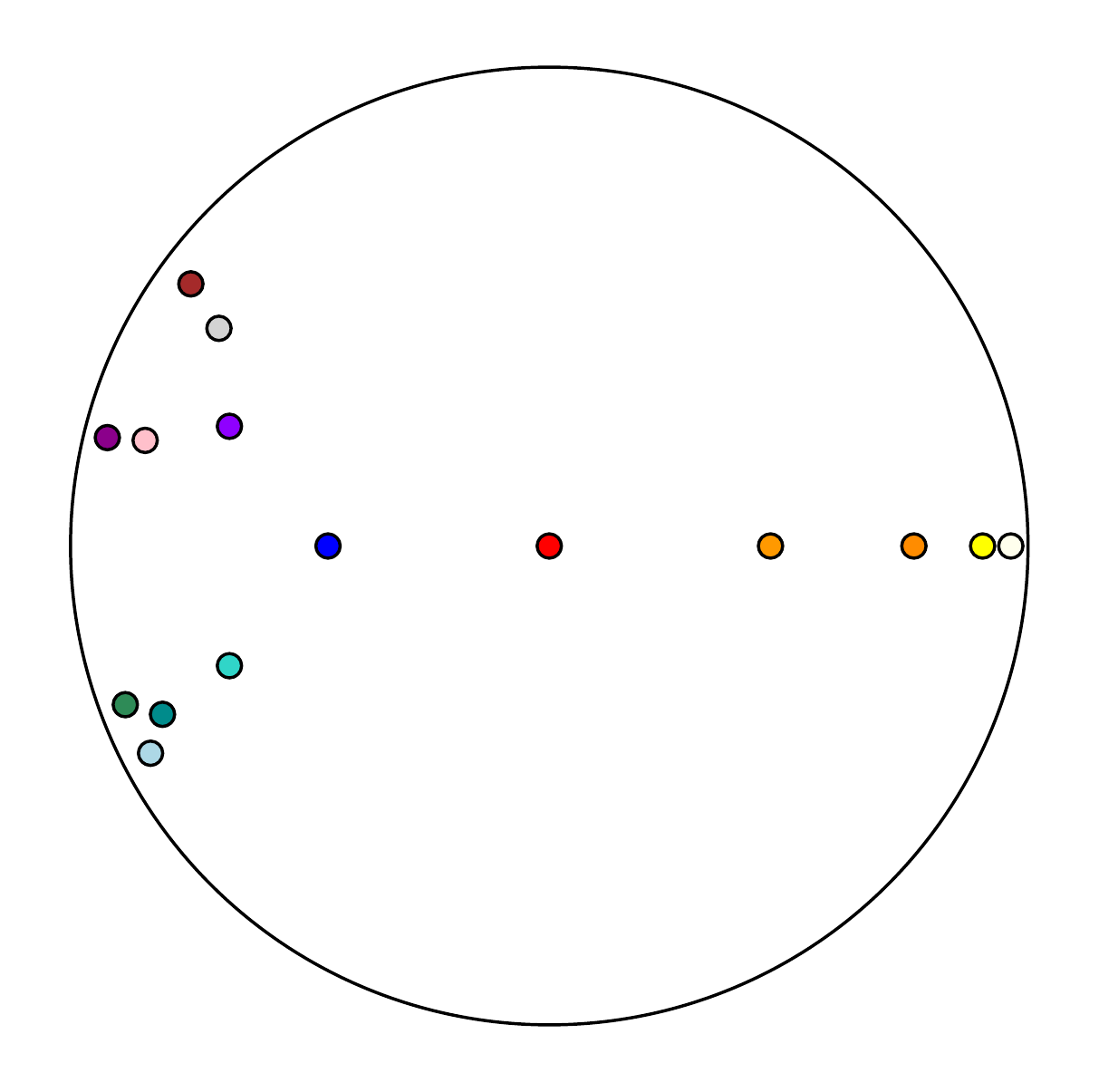}
    \captionof{figure}{Embeddings of the nodes of the hierarchy of Figure~\ref{fig:example} into the Poincar{\'e} disk $\mathbb{D}^2$ generated by Sarkar's construction.}
    \label{fig:sarkar}
\end{minipage}
\end{figure}

\section{Sarkar's Construction}\label{sec:sarkar}
There exists a construction proposed by Sarkar which can embed the nodes of a tree into the Poincar{\'e} disk $\mathbb{D}^2$ and preserve the distances between nodes with arbitrarily low distortion~\citep{sarkar2011low}.
Sarkar's construction, shown in Algorithm~\ref{alg:sarkar}, is purely combinatorial.
The inputs to the Algorithm are a scaling factor $\tau$, a node $a$ (of degree $\text{deg}(a)$) from the tree and its parent node $b$.
These nodes have already been embedded into $\mathbb{D}^2$, and the objective is to to also embed the $\text{deg}(a) - 1$ children of $a$ into that space.
Specifically, to place the children $c_1, c_2, \ldots, c_{\text{deg}(a)-1}$ of node $a$ into $\mathbb{D}^2$, the algorithm performs the following steps: ($1$) it reflects the embeddings of nodes $a$ and $b$ ($f(a)$ and $f(b)$, respectively) across a geodesic such that $f(a)$ is mapped onto the origin (\ie $0$) and $f(b)$ is mapped onto some point $z$; ($2$) it places the children of node $a$ to vectors $y_1,y_2,\ldots,y_{\text{deg}(a)-1}$ equally spaced around a circle with radius $\nicefrac{\exp(\tau) - 1}{\exp(\tau) + 1}$ (which is a circle of radius $\tau$ in the hyperbolic metric), and maximally separated from node $b$'s reflected  embedding $z$; and ($3$) it reflects all of the points back across the geodesic, thus the origin is mapped back onto $f(a)$, while $z$ is mapped onto $f(b)$.
To embed the entire tree, the algorithm follows a recursive approach: the root is placed at the origin and its children in a circle around it, and then the children of all nodes are embedded in the space until no more nodes are left to be placed.
This complexity of the entire algorithm in linear in the number of nodes of the tree.
Figure~\ref{fig:sarkar} illustrates how Sarkar's construction embeds the nodes of the hierarchy of Figure~\ref{fig:example} into the Poincar{\'e} disk $\mathbb{D}^2$.

One limitation of Sarkar's construction is that the number of bits of precision used to represent components of the embedded nodes scales linearly with the maximum path length~\citep{sala2018representation}.
Thus, Sarkar's construction might experience numerical instabilities in the case of trees that contain long paths.
\cite{sala2018representation} generalized Sarkar's construction from the Poincare disk $\mathbb{D}^2$ to the Poincare ball $\mathbb{D}^d$ to deal with such problems.
The new construction can produce embeddings of higher quality in the case of bushy trees, \ie trees whose maximum degree is large.

\section{Additional Experiments}\label{sec:add_experiments}
\subsection{Comparison to WL Distance}
\begin{figure}
    \begin{subfigure}{.33\textwidth}
      \centering
      \includegraphics[width=\linewidth]{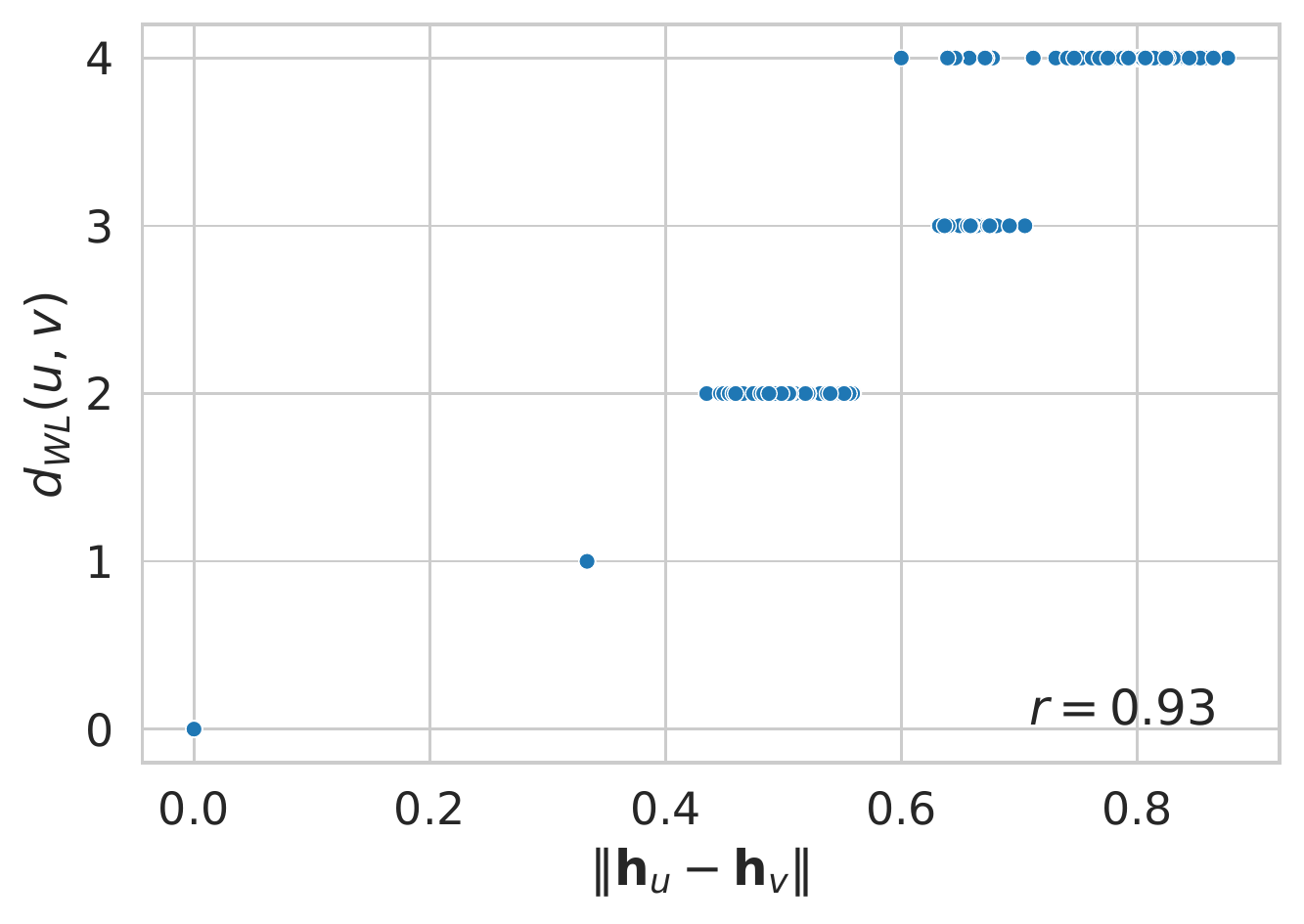}
      \caption{Algorithm of~\cite{sala2018representation}}
    \end{subfigure}%
    \begin{subfigure}{.31\textwidth}
      \centering
      \includegraphics[width=\linewidth]{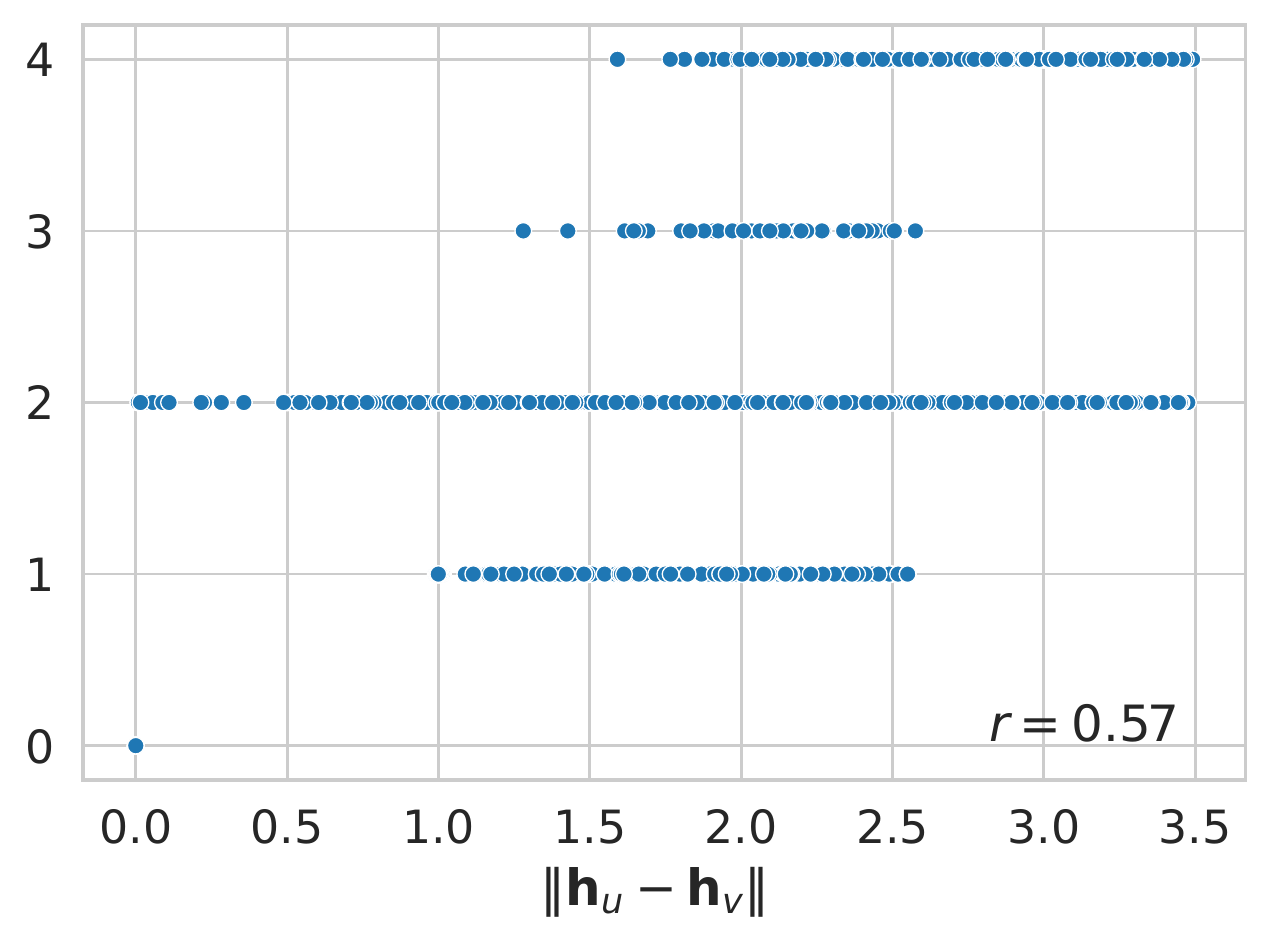}
      \caption{WLHN model}
    \end{subfigure}%
    \begin{subfigure}{.31\textwidth}
      \centering
      \includegraphics[width=\linewidth]{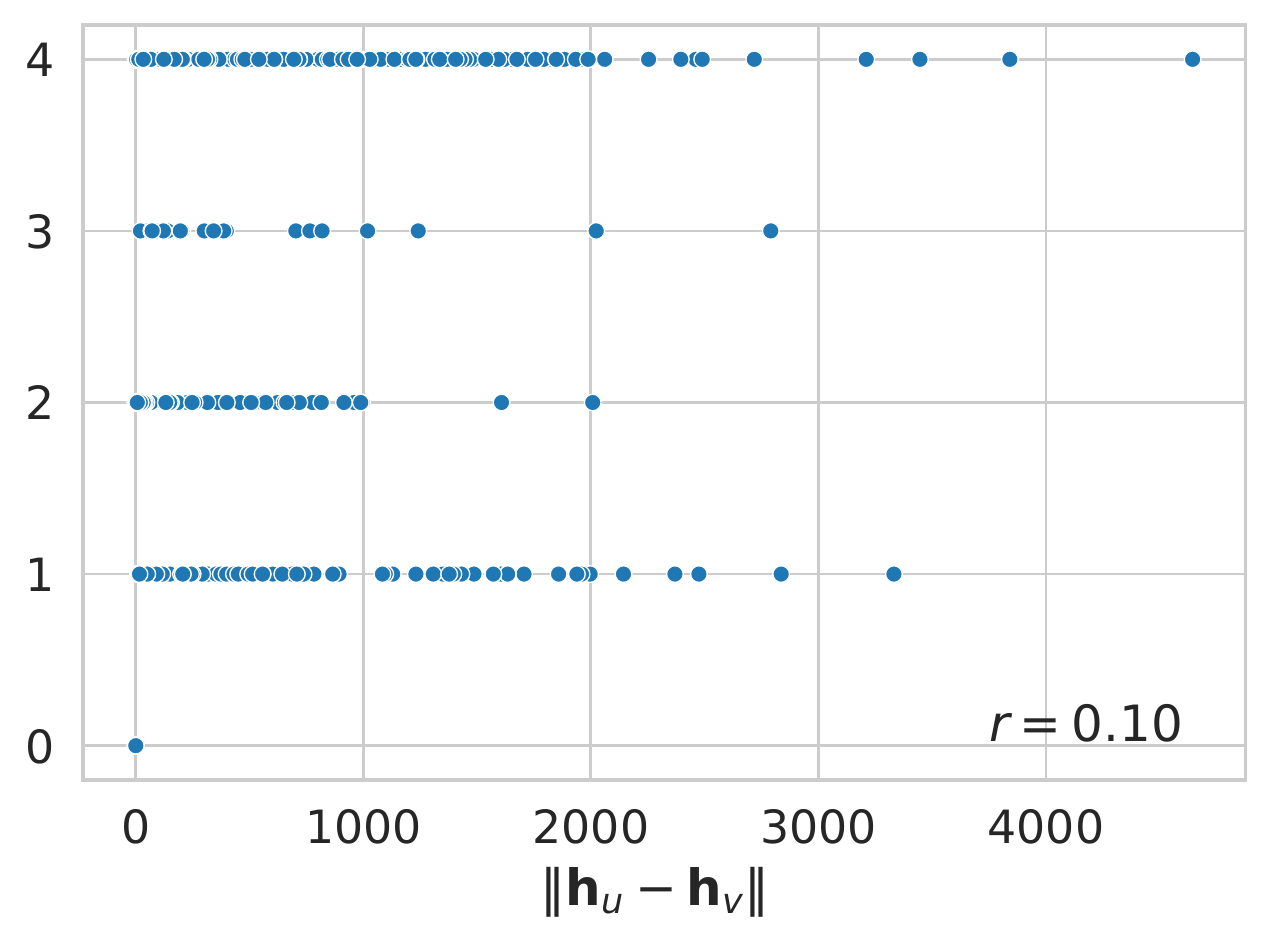}
      \caption{GIN model}
    \end{subfigure}
    \caption{WL distances of the nodes of the the hierarchy produced by the WL algorithm on the IMDB-BINARY dataset vs. distances of the corresponding embeddings of the nodes produced by the algorithm of~\cite{sala2018representation}, the proposed WLHN model and the GIN model.}
    \label{fig:distances}
\end{figure}

We performed an experiment where we extracted the hierarchy produced by the WL algorithm on the IMDB-BINARY dataset, and then compared the WL distances of the nodes of the hierarchy against the distances of the embeddings of the nodes produced by the algorithm of~\cite{sala2018representation}, by the WLHN model and by the GIN model.
All nodes are initially annotated with a single feature equal to $1$.
We set the number of iterations of the WL algorithm to $2$.
Thus, the WL distances are at most equal to $4$.
We set the dimension of the embeddings to $128$ for all approaches.
We did not train the WLHN and GIN models, but we randomly initialized their parameters and performed a feed-forward pass to obtain the embeddings.
There are $2,997$ nodes in the hierarchy, thus we compute $2,997^2$ distances in total.
We visualize those distances in Figure~\ref{fig:distances} along with the Pearson correlation coefficients.
As expected, the algorithm of~\cite{sala2018representation} achieves the highest value of correlation, followed by the proposed WLHN model.
GIN achieves the lowest value of correlation.
Specifically, the three methods achieve correlations equal to $0.93, 0.57$ and $0.10$, respectively.
Note that the objective of the proposed model is not to achieve a very high value of correlation (\ie very close to $1$).
Consider the following example: the nodes of the hierarchy that are directly connected to the root represent the different degrees of the nodes of all graphs of the IMDB-BINARY dataset.
The WL distance between any two of these nodes is equal to $2$.
However, in practice, we would like the node that represents a value of degree equal to $1$ to be closer to the node that represents a value of degree equal to $3$ than to the node that represents a value of degree equal to $100$.
The proposed model can achieve that since it takes the nodes' Euclidean embeddings into account which capture such relationships between nodes.

\subsection{Running Time}
In Table~\ref{tab:running_time_results}, we present the average running time per epoch of the proposed model and two baselines (GIN~\citep{xu2019powerful} and HGCN~\citep{chami2019hyperbolic}) on the $10$ graph classification datasets (experiments performed on an Nvidia Titan Xp GPU).
As discussed in section~\ref{sec:contribution}, the time complexity of the model is in the same order of magnitude as those of standard MPNNs.
We observe that the running time of WLHN is comparable with those of GIN and HGCN and thus not prohibitive for real-world problems.

\begin{table}[t]
\caption{Average running time per epoch (in seconds).}
\label{tab:running_time_results}
\centering
\scriptsize
\addtolength{\tabcolsep}{-2.8pt}
\def\arraystretch{1.1}
\begin{tabular}{lcccccccccc}
\toprule
& \textbf{MUTAG} & \textbf{D\&D} & \textbf{NCI1} & \textbf{PROTEINS} & \textbf{ENZYMES} & \textbf{IMDB-B} & \textbf{IMDB-M} & \textbf{REDDIT-B} & \textbf{REDDIT-5K} & \textbf{COLLAB} \\
\midrule
GIN & 0.09  & 5.51 & 0.73 & 0.24  &  0.14 & 0.22 & 0.34 & 0.77 & 2.01 & 15.30  \\ 
HGCN & \multirow{2}{*}{0.13} & \multirow{2}{*}{6.58} & \multirow{2}{*}{1.32}  & \multirow{2}{*}{0.44}  & \multirow{2}{*}{0.23} & \multirow{2}{*}{0.43} & \multirow{2}{*}{0.63} & \multirow{2}{*}{2.10} & \multirow{2}{*}{7.10} & \multirow{2}{*}{16.20} \\
(PoincareBall) \\
\midrule
\textbf{WLHN} & 0.18  & 5.85  & 2.07 & 0.61  & 0.16 & 0.50 & 0.73 & 2.30  & 6.83 & 16.55 \\
\bottomrule
\end{tabular}
\addtolength{\tabcolsep}{1pt}
\end{table}

\begin{table}[t]
        \centering
        \def\arraystretch{1.1}
        \caption{Classification accuracy ($\pm$ standard deviation) on node classification tasks with feature noise.}
        \label{tab:rob_features}
        \resizebox{\textwidth}{!}{
        \begin{tabular}{l|ccc|ccc|ccc}
        \toprule 
        \multirow{2}{*}{\textbf{Method}} &
        \multicolumn{3}{c}{\textbf{Cornell}}& \multicolumn{3}{c}{\textbf{Texas}} & \multicolumn{3}{c}{\textbf{Wisconsin}} \\ &
        \multirow{2}{*}{Without Noise} & Feature Noise & Feature Noise &  \multirow{2}{*}{Without Noise} & Feature Noise & Feature Noise &  \multirow{2}{*}{Without Noise} & Feature Noise & Feature Noise
        \\
        & & (0.1) & (0.2) &  & (0.1) & (0.2) & & (0.1) & (0.2) \\
        \midrule          
         GCN & 52.16 ($\pm$ 8.20) & 48.10 ($\pm$ 1.16) & 50.13 ($\pm$ 7.64) & 56.49 ($\pm$ 8.83) & 54.05 ($\pm$ 8.29) & 55.67 ($\pm$ 6.41) & 48.43 ($\pm$ 4.38) & 48.24 ($\pm$ 4.65) & 49.80 ($\pm$ 7.08) \\   
        GIN & 51.62 ($\pm$ 7.68) & 49.73 ($\pm$ 9.83) & 50.81 ($\pm$ 9.80) & 53.24 ($\pm$ 8.55) & 50.81 ($\pm$ 9.80) & 50.15 ($\pm$ 9.34) & 50.59 ($\pm$ 7.98)  & 51.17 ($\pm$ 9.17) & 48.23 ($\pm$ 6.86) \\ 
        HGCN & \multirow{2}{*}{64.59 ($\pm$ 11.10)} & \multirow{2}{*}{61.89 ($\pm$ 7.30)} & \multirow{2}{*}{58.92 ($\pm$ 8.27 )} & \multirow{2}{*}{61.62 ($\pm$ 9.34)} & \multirow{2}{*}{ 60.00 ($\pm$ 8.35 )} & \multirow{2}{*}{58.65 ($\pm$ 7.05 )} & \multirow{2}{*}{70.20 ($\pm$ 7.73)} & \multirow{2}{*}{68.23 ($\pm$ 9.71)} & \multirow{2}{*}{67.53 ($\pm$ 10.46)} \\
        (Hyperboloid) \\
        \midrule
        \textbf{WLHN} & \textbf{75.41} ($\pm$ 6.67) & \textbf{67.83} ($\pm$ 7.78) & \textbf{67.81} ($\pm$ 7.33) & \textbf{74.59} ($\pm$ 7.60) & \textbf{70.54} ($\pm$ 8.23) & \textbf{67.02} ($\pm$ 8.00) & \textbf{77.65} ($\pm$ 5.56) & \textbf{75.88} ($\pm$ 5.48 )& \textbf{77.23} ($\pm$ 4.59) \\
        \bottomrule
        \end{tabular}}
\end{table}

\begin{table}[t]
        \centering
        \def\arraystretch{1}
        \caption{Classification accuracy ($\pm$ standard deviation) on node classification tasks with structural noise.}
        \label{tab:rob_structural}
        \resizebox{\textwidth}{!}{
        \begin{tabular}{l|ccc|ccc|ccc}
        \toprule 
        \multirow{2}{*}{\textbf{Method}} &
        \multicolumn{3}{c}{\textbf{Cornell}}& \multicolumn{3}{c}{\textbf{Texas}} & \multicolumn{3}{c}{\textbf{Wisconsin}} \\ &
        \multirow{2}{*}{Without Noise} & Structural Noise & Structural Noise & \multirow{2}{*}{Without Noise} & Structural Noise & Structural Noise & \multirow{2}{*}{Without Noise} & Structural Noise & Structural Noise \\
        & & (0.1) & (0.2) & & (0.1) & (0.2) & & (0.1) & (0.2) \\
        \midrule          
        HGCN & \multirow{2}{*}{64.59 ($\pm$ 11.10)} & \multirow{2}{*}{ 58.91 ($\pm$ 10.31)} & \multirow{2}{*}{ 54.86 ($\pm$ 9.97)} & \multirow{2}{*}{61.62 ($\pm$ 9.34)} & \multirow{2}{*}{ 57.03 ($\pm$ 8.33)} & \multirow{2}{*}{58.65 ($\pm$ 9.05 )} & \multirow{2}{*}{70.20 ($\pm$ 7.73)} & \multirow{2}{*}{69.76 ($\pm$ 7.95)} & \multirow{2}{*}{ 67.32 ($\pm$ 8.47)} \\
        (Hyperboloid) \\
        \midrule
        \textbf{WLHN} & \textbf{75.41} ($\pm$ 6.67) & \textbf{72.70} ($\pm$ 6.78) & \textbf{68.65} ($\pm$ 8.12) & \textbf{74.59} ($\pm$ 7.60) & \textbf{71.08} ($\pm$ 6.94) & \textbf{74.05} ($\pm$ 5.94) & \textbf{77.65} ($\pm$ 5.56) & \textbf{77.45} ($\pm$ 6.39)& \textbf{77.04} ($\pm$ 5.02) \\
        \bottomrule
        \end{tabular}}
\end{table}

\begin{figure}[t]
    \centering
    \includegraphics[width=0.7\textwidth]{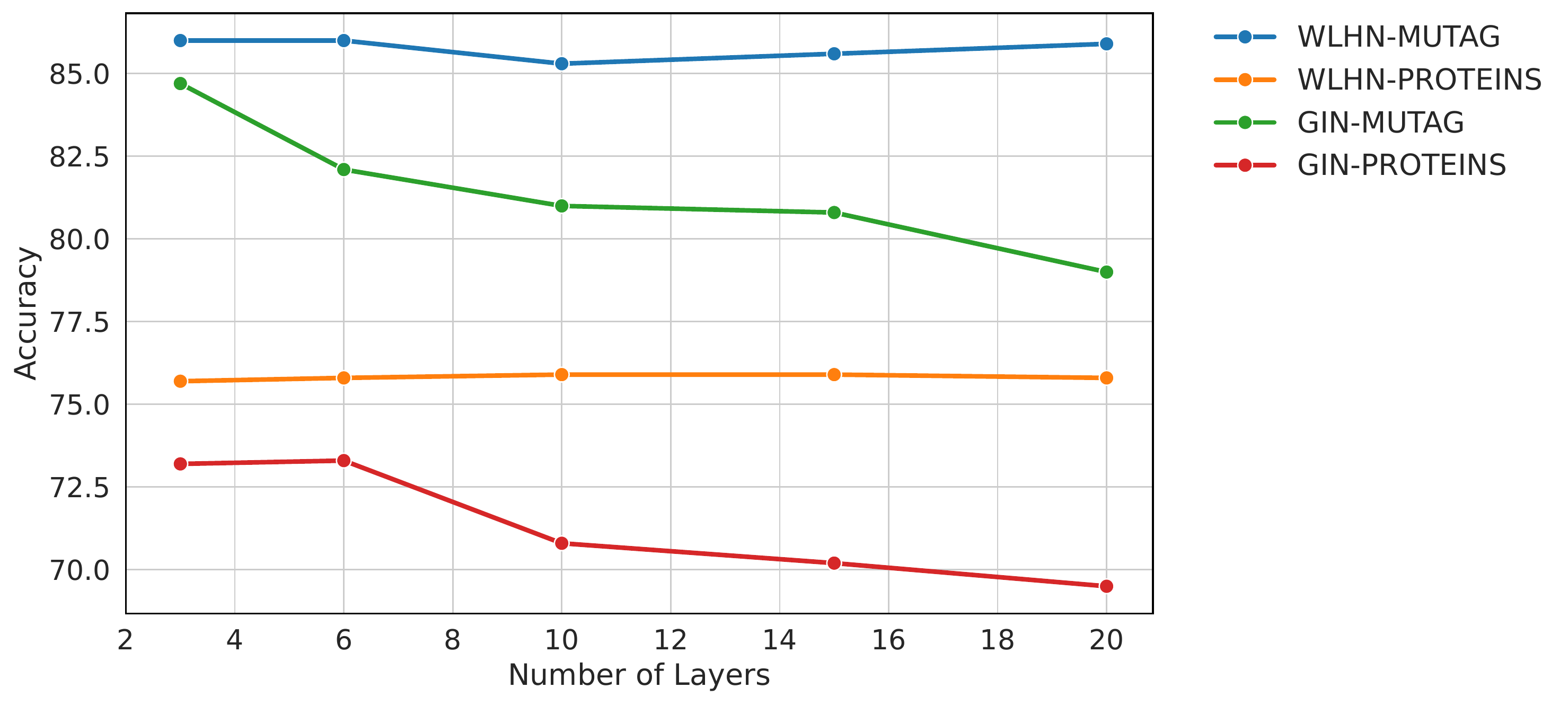} 
    \caption{Test accuracy of WLHN and GIN with respect to the numeber of layers}
    \label{fig:oversmoothing}
\end{figure}

\subsection{Robustness Experiments}
We have performed experiments investigating the performance of our model after applying different types of perturbations on the input.
Specifically, we use the Cornell, Texas, Wisconsin datasets and we apply Gaussian noise with $\mu=2$ and $\sigma=0.2$ to the node features of the training graphs.
We report in Table \ref{tab:rob_features} the mean classification accuracy and the standard deviation when $10\%$ or $20\%$ of the nodes are affected by the noise, across $10$ different splits.
We observe that noise has a slight impact on the performance of all models, while the proposed WLHN model still significantly outperforms the baselines on all $3$ datasets.

We further investigate the performance of the WLHN model after applying structural noise.
Specifically, we randomly sample nodes and we add $k$ edges between nodes associated with different class labels. 
We report the results in Table \ref{tab:rob_structural} for $k \approx 0.1 |E|$ and $k \approx 0.2|E|$, where $|E|$ is the number of the edges of the graph, across $10$ different random splits. 
We compare the performance of WLHN against that of HGCN (Hyperboloid).
We observe that structural noise also does not have a large impact on the performance of the proposed model, and that WLHN is more robust to noise than HGCN (Hyperboloid).
These results partially validate our intuition that the proposed model is more robust to noise than existing models that do not explicitly capture the nodes' structural distance in the generated representations. 

\subsection{Oversmoothing Experiments}
From the theoretical point of view, it is known from the proposed construction that the distance between two nodes in the hyperbolic space cannot decrease as more neighborhood aggregation steps are performed. Thus, over-smoothing cannot occur, i.e., the node representations of structurally different nodes (according to WL) will never converge to indistinguishable vectors.
We also experimentally demonstrate that our model can have increased depth without suffering from over-smoothing.
We performed experiments on the MUTAG and PROTEINS datasets and report the results in Figure~\ref{fig:oversmoothing}.
We observe that even with a large number of layers, there is no decrease in the performance of our model.

\section{Hyperparameters}\label{sec:hyperparams}
For a fair comparison, we used the same hyperparameters setup between our model and HGCN\citep{chami2019hyperbolic}.
For all datasets, the hidden dimension size of the GNN layers was chosen from $\{ 32, 64, 128\}$. 
The number of neighborhood aggregation layers was chosen from $\{1,\ldots,5\}$ and we used batch normalization.
The emerging graph representations were fed to a multi-layer perceptron consisting of two hidden layers with a hidden-dimension size of $128$ and $64$, respectively.
We used the ReLU as an activation function.  
We used dropout with a ratio chosen from $\{p=0.0, 0.5\}$ between the hidden layers.
We trained the model by minimizing the cross-entropy loss.
We choose between two optimizers, Adam and RiemannianAdam, with a learning rate in $\{10^{-3},10^{-2}\}$.
We trained the networks for $300$ epochs.
The experiments were run on a Intel(R) Xeon(R) CPU E5-1607 v2 @ 3.00GHz processor with $128$GB ram and an Nvidia TITAN Xp GPU.

\begin{table}[t]
  \caption{Summary of the $10$ datasets that were used in our experiments for graph classification.}
  \label{tab:statistics}
  \begin{center}
  \resizebox{\textwidth}{!}{%
  \begin{tabular}{lcccccccccc}
  \toprule
  \multirow{2}{*}{\textbf{Dataset}} & \multirow{2}{*}{MUTAG} & \multirow{2}{*}{D\&D} & \multirow{2}{*}{NCI1} & \multirow{2}{*}{PROTEINS} & \multirow{2}{*}{ENZYMES} & IMDB & IMDB & REDDIT & REDDIT & \multirow{2}{*}{COLLAB} \\ 
  & & & & & & BINARY & MULTI & BINARY & MULTI-5K &  \\
  \midrule
  Max \# vertices & 28 & 5,748 & 111 & 620 & 126 & 136 & 89 & 3,782 & 3,648 & 492 \\
  Min \# vertices & 10 & 30 & 3 & 4 & 2 & 12 & 7 & 6 & 22 & 32 \\
  Average \# vertices & 17.93 & 284.32 & 29.87 & 39.05 & 32.63 & 19.77 & 13.00 & 429.61 & 508.50 & 74.49 \\ 
  Max \# edges & 33 & 14,267 & 119 & 1,049 & 149 & 1,249 & 1,467 & 4,071 & 4,783 & 40,119 \\
  Min \# edges & 10 & 63 & 2 & 5 & 1 & 26 & 12 & 4 & 21 & 60 \\
  Average \# edges & 19.79 & 715.66 & 32.30 & 72.81 & 62.14 & 96.53 & 65.93 & 497.75 & 594.87 & 2,457.34 \\ 
  \# labels & 7 & 82 & 37 & 3 & -- & -- & -- & -- & -- & -- \\ 
  \# attributes & -- & -- & -- & -- & 18 & -- & -- & -- & -- & -- \\ 
  \# graphs & 188 & 1,178 & 4,110 & 1,113 & 600 & 1,000 & 1,500 & 2,000 & 4,999 & 5,000 \\ 
  \# classes & 2 & 2 & 2 & 2 & 6 & 2 & 3 & 2 & 5 & 3 \\
  \bottomrule
  \end{tabular}
  }
  \end{center}
  \vskip -0.1in
\end{table} 

\begin{table}[t]
  \begin{center}
    \caption{Dataset statistics and properties for node-level prediction tasks.}
    \resizebox{0.6\textwidth}{!}{   \renewcommand{\arraystretch}{1.05}
      \begin{tabular}{@{}lccc@{}}\toprule
      \multirow{3}{*}{\vspace*{4pt}\textbf{Dataset}}&\multicolumn{3}{c}{\textbf{Properties}}\\
        \cmidrule{2-4} & Number of nodes & Number of edges & Number of node features \\ \midrule
        $\textsc{Cornell}$   & 183             & 295 & 1,703 \\
        $\textsc{Texas}$ & 183 & 309             & 1,703 \\
        $\textsc{Wisconsin}$ & 251             & 490 & 1,703 \\
        $\textsc{Squirrel}$ & 5,201             & 217,073 & 2,089 \\
        $\textsc{Actor}$ & 7,600             & 33,544 & 931 \\

        \bottomrule
      \end{tabular}}
    \label{tab:dss}
  \end{center}
\end{table}

\section{Datasets}\label{sec:datasets}
We evaluated the proposed model on $10$ publicly available graph classification datasets including $5$ bio/chemo-informatics datasets: MUTAG, D\&D, NCI1, PROTEINS and ENZYMES, as well as $5$ social interaction datasets: IMDB-BINARY, IMDB-MULTI, REDDIT-BINARY, REDDIT-MULTI-5K and COLLAB \citep{morris2020tudataset}.
A summary of the $10$ datasets is given in Table~\ref{tab:statistics}.
MUTAG consists of $188$ mutagenic aromatic and heteroaromatic nitro compounds.
The task is to predict whether or not each chemical compound has a mutagenic effect on the Gram-negative bacterium {\it Salmonella typhimurium} \citep{debnath1991structure}.
ENZYMES contains $600$ protein tertiary structures represented as graphs obtained from the BRENDA enzyme database.
Each enzyme is a member of one of the Enzyme Commission top level enzyme classes (EC classes) and the task is to correctly assign the enzymes to their classes \citep{borgwardt2005protein}.
NCI1 contains more than four thousand chemical compounds screened for activity against non-small cell lung cancer and ovarian cancer cell lines \citep{wale2008comparison}.
PROTEINS contains proteins represented as graphs where vertices are secondary structure elements and there is an edge between two vertices if they are neighbors in the amino-acid sequence or in $3$D space.
The task is to classify proteins into enzymes and non-enzymes \citep{borgwardt2005protein}. 
D\&D contains over a thousand protein structures.
Each protein is a graph whose nodes correspond to amino acids and a pair of amino acids are connected by an edge if they are less than $6$ \AA angstroms apart.
The task is to predict if a protein is an enzyme or not \citep{dobson2003distinguishing}.
IMDB-BINARY and IMDB-MULTI were created from IMDb, an online database of information related to movies and television programs. 
The graphs contained in the two datasets correspond to movie collaborations.
The vertices of each graph represent actors/actresses and two vertices are connected by an edge if the corresponding actors/actresses appear in the same movie.
Each graph is the ego-network of an actor/actress, and the task is to predict which genre an ego-network belongs to \citep{yanardag2015deep}.
REDDIT-BINARY and REDDIT-MULTI-5K contain graphs that model the social interactions between users of Reddit.
Each graph represents an online discussion thread.
Specifically, each vertex corresponds to a user, and two users are connected by an edge if one of them responded to at least one of the other's comments.
The task is to classify graphs into either communities or subreddits \citep{yanardag2015deep}.
COLLAB is a scientific collaboration dataset that consists of the ego-networks of several researchers from three subfields of Physics (High Energy Physics, Condensed Matter Physics and Astro Physics).
The task is to determine the subfield of Physics to which the ego-network of each researcher belongs \citep{yanardag2015deep}.

We also evaluated the proposed kernel on two datasets from the Open Graph Benchmark (OGB) \citep{hu2020open}, a collection of large-scale and diverse benchmark datasets for machine learning on graphs.
They are both molecular property prediction datasets that are adopted from the MoleculeNet \citep{wu2018moleculenet}.
\texttt{ogbg-molhiv} consists of $41,127$ molecules and corresponds to a binary classification dataset where the task is to predict whether a molecule inhibits HIV virus replication or not.
The average number of vertices per graph is equal to $25.5$, while the average number of edges is equal to $27.5$.
The dataset is split into training/validation/test sets with a ratio of $80/10/10$.
For \texttt{ogbg-molpcba} dataset, the class balance is very skewed (only 1.4\% of data is positive) and the dataset contains $128$ classification tasks. Therefore, we use the Average Precision (AP) averaged over the tasks as the evaluation metric.
It contains $437,929$ molecule graphs with an average number of vertices per graph equal to $26.0$, and an average number of edges equal to $28.1$.
For both datasets, the molecules in the training, validation, and test sets are divided using a scaffold splitting procedure that splits the molecules based on their two-dimensional structural frameworks.
The scaffold splitting attempts to separate structurally different molecules into different subsets.

Finally, we evaluated the proposed model on five node classification datasets: Cornell, Texas, Wisconsin, Squirrel, and Actor.
The first three datasets are extracted from the WebKB dataset and the rest of the datasets from Wikipedia, and have been employed in previous studies~\citep{pei2020geom,rozemberczki2021multi}.
A summary of the three datasets is given in Table~\ref{tab:dss}.

\vfill

\end{document}